\documentclass{article}

\PassOptionsToPackage{numbers,sort&compress}{natbib}
 \usepackage{amsmath}
\usepackage[eandd,final]{neurips_2026}

\usepackage[T1]{fontenc}
\usepackage[utf8]{inputenc}
\usepackage{microtype}
\usepackage{booktabs}
\usepackage{multirow}
\usepackage{amsmath,amssymb,amsfonts}
\usepackage{mathtools}
\usepackage{graphicx}
\usepackage{subcaption}
\usepackage{url}
\usepackage{enumitem}
\usepackage{xcolor}
\usepackage{hyperref}
\usepackage{booktabs}
\usepackage{xcolor}
\usepackage{pifont}
\usepackage[numbers]{natbib}
\newcommand{\cmark}{\ding{51}}
\newcommand{\xmark}{\ding{55}}

\newcommand{\method}{\textsc{SpatialEpiBench}}

\title{\method{}: Benchmarking Spatial Information and Epidemic Priors in Forecasting}

\author{
Ruiqi Lyu$^{*,1}$ \quad
Alistair Turcan$^{*,1}$ \quad
Bryan Wilder$^{1}$\\
$^{1}$Carnegie Mellon University\\
\texttt{\{ruiqil, aturcan, bwilder\}@cs.cmu.edu}\\
$^*$Equal contribution.
}

\begin{document}

\maketitle

\begin{abstract}
Accurate epidemic forecasting is crucial for public health response, resource allocation, and outbreak intervention, but remains difficult with sparse, noisy, and highly non-stationary data.
Because epidemics unfold across interacting regions, spatiotemporal methods are natural candidates for improving forecasts.
Despite growing interest in spatial information, recent spatiotemporal epidemic forecasting studies use heterogeneous datasets and protocols, often with a single chronological split.
We address this gap with \method{}, a standardized benchmark for retrospective rolling-origin evaluation of spatiotemporal epidemic forecasting.
\method{} includes 11 primary epidemic datasets with standardized rolling evaluations and outbreak-specific metrics.
We evaluate adjacency-informed forecasting models with widely used epidemic priors that adapt general models to epidemiology, but find that most methods underperform a simple last-value baseline from 1 day to 1 month ahead, even during outbreaks and with these priors.
We identify three major failure modes: (1) poor outbreak anticipation, (2) difficulty handling sparsity and noise, and (3) limited utility of common geographic adjacency for epidemiological spatial information.
We release benchmark data, code, and instructions at
\url{https://github.com/Rachel-Lyu/SpatialEpiBench}
to support development of operationally useful epidemic forecasting models.
\end{abstract}

\section{Introduction}
Epidemic forecasting is essential for public health planning, guiding resource allocation and preventive interventions before outbreaks occur \citep{cramer2022pnas, rivers2019using}.
Yet surveillance data are sparse, noisy, and non-stationary, shaped by reporting artifacts, behavioral shifts, policy changes, pathogen evolution, and seasonality \citep{parag2022quantifying, rodriguez2024machine}, making them difficult to forecast.
Spatiotemporal models exploit dependencies across related or interacting regions and have performed well in traffic and energy forecasting \citep{bai2020adaptive, li2017diffusion, wu2020connecting}.
They are therefore natural candidates for epidemic forecasting, where outbreaks may spread across related populations and signals from other regions may provide early warnings.
A growing literature adapts these models to epidemiology by adding epidemic priors or architectural components that capture infectious-disease mechanisms and epidemic-specific patterns \citep{liu2025review}.
However, spatiotemporal epidemic forecasting still lacks an independent standardized benchmark.
Existing spatiotemporal epidemic evaluations use heterogeneous datasets and protocols, often omitting repeated rolling evaluations and outbreak-specific forecasts (Table~\ref{tab:related_benchmarks}). Rolling-origin evaluation is established practice \citep{hyndman2021fpp3}; our contribution is to harmonize it with common targets, horizons, baselines, and diagnostics.
Without unified, practically relevant evaluation, it remains difficult to determine whether spatial information and epidemic priors genuinely improve forecasting utility.


To address this gap, we propose \textbf{\method{}}, a standardized epidemic forecasting benchmark. We use this benchmark to study in detail the utility of epidemic priors and spatial information in forecasting.
\method{} provides a standardized retraining and rolling evaluation procedure for 9 daily and 2 weekly spatiotemporal epidemic datasets, comparing against 3 univariate baselines across multiple horizons, modalities, and 9 metrics, with emphasis on outbreak periods.
It also includes 4 model-agnostic epidemic priors applicable to any deep learning model, enabling tests of whether domain-specific structure improves general architectures without bespoke redesign.
To our knowledge, \method{} is the most comprehensive spatiotemporal epidemic forecasting benchmark to date and the first to jointly study spatial information and epidemic adaptation of general forecasters.

In this work, we show that when using a common geographic adjacency as input, many spatiotemporal forecasting methods underperform simply predicting the last observed value.
This holds from 1 day to 1 month ahead, even during outbreaks, when such a baseline should fail.
A systematic search over epidemic priors shows that some methods improve when epidemic patterns are explicitly modeled, but adjacency-informed methods do not consistently outperform simpler univariate baselines.
We diagnose three sources of underperformance: existing methods under-predict outbreaks, struggle with zeros and outliers, and commonly used geographic adjacency may provide weak epidemiological spatial information.

Our contributions are as follows:
\begin{enumerate}
    \item We present a large, standardized benchmark of 11 spatiotemporal epidemic forecasting tasks, with unified data handling, training protocols, and evaluation procedures, including a special focus on outbreak periods.
    \item We evaluate 11 spatiotemporal forecasting methods with geographic adjacency against 3 univariate baselines, finding that many models often fail to translate this spatial structure into strong performance, even with the added help of epidemic priors.
    \item We do a deep dive into why models fail, finding outbreaks are often under predicted or missed entirely, sparsity and outliers substantially hinder performance, and geographic adjacency does not generally provide useful information for forecasting.
\end{enumerate}

\begin{table}[hbt!]
\centering
\small
\setlength{\tabcolsep}{3.5pt}
\begin{tabular}{llcccccc}
\toprule
\shortstack{\textbf{Type}\\\textbf{}} &
\shortstack{\textbf{Benchmark/}\\\textbf{Method}} &
\shortstack{\textbf{Epi}\\\textbf{datasets}} &
\shortstack{\textbf{Spatio-}\\\textbf{temporal}} &
\shortstack{\textbf{Rolling}\\\textbf{evals}} &
\shortstack{\textbf{Outbreak}\\\textbf{focused}} &
\shortstack{\textbf{Evaluates}\\\textbf{priors}} &
\shortstack{\textbf{Metrics}\\\textbf{}} \\
\midrule
Benchmark & Dey et al \citep{dey2024we} & 15 & \textcolor{red}{\xmark} & \textcolor{green!60!black}{\cmark} & \textcolor{red}{\xmark} & \textcolor{red}{\xmark} & 4 \\
 & Panja et al \citep{panja2025zero} & 11 & \textcolor{red}{\xmark} & \textcolor{green!60!black}{\cmark} & \textcolor{red}{\xmark} & \textcolor{red}{\xmark} & 4 \\
 & IDOBE \citep{adiga2026idobe} & 13 & \textcolor{red}{\xmark} & \textcolor{green!60!black}{\cmark} & \textcolor{green!60!black}{\cmark} & \textcolor{red}{\xmark} & 3 \\
\midrule
Method & Cola-GNN \citep{deng2020cola} & 2 & \textcolor{green!60!black}{\cmark} & \textcolor{red}{\xmark} & \textcolor{red}{\xmark} & \textcolor{red}{\xmark} & 3 \\
 & EARTH \citep{wan2025earth} & 3 & \textcolor{green!60!black}{\cmark} & \textcolor{red}{\xmark} & \textcolor{green!60!black}{\cmark} & \textcolor{red}{\xmark} & 2 \\
 & EpiGNN \citep{xie2022epignn} & 5 & \textcolor{green!60!black}{\cmark} & \textcolor{red}{\xmark} & \textcolor{red}{\xmark} & \textcolor{red}{\xmark} & 2 \\
 & EpiColaGNN \citep{liu2023epidemiology} & 1 & \textcolor{green!60!black}{\cmark} & \textcolor{red}{\xmark} & \textcolor{red}{\xmark} & \textcolor{red}{\xmark} & 3 \\
\midrule
This paper & \textbf{\method{}} & 11 & \textcolor{green!60!black}{\cmark} & \textcolor{green!60!black}{\cmark} & \textcolor{green!60!black}{\cmark} & \textcolor{green!60!black}{\cmark} & 9 \\
\bottomrule
\end{tabular}
\caption{\method{} characteristics versus a representative set of related benchmarks. 
}
\label{tab:related_benchmarks}
\end{table}

\section{Related work}

\paragraph{Epidemic forecasting}

Accurate forecasts of cases, deaths, hospitalizations, and other operational targets are central to public health planning and intervention \citep{emanuel2020fair}.
Mechanistic models, including compartmental ODE systems such as SIR \citep{hethcote2000mathematics}, are interpretable but often rely on simplified assumptions that are difficult to reconcile with real-world dynamics \citep{dehning2020inferring}.
Statistical and machine learning methods such as ARIMA \citep{alghamdi2019forecasting} and DLinear \citep{zeng2023transformers} have shown strong univariate time-series performance, including in epidemiology \citep{dey2024we, panja2025zero}.
Epidemic forecasting remains difficult because surveillance data are sparse, noisy, and non-stationary, with reporting artifacts, seasonality, behavioral shifts, holidays, and policy changes creating unstable predictive patterns \citep{parag2022quantifying, rodriguez2024machine}.
These challenges have motivated epidemiology-specific forecasting methods and benchmarks \citep{dey2024we, panja2025zero}, as well as large collaborative efforts such as the CDC-associated COVID-19 Forecast Hub, which aggregated forecasts from 110 models \citep{cramer2022united}.
Benchmarks such as IDOBE \citep{adiga2026idobe} now measure univariate outbreak forecasting in realistic settings.
Because epidemics spread and interact regions, recent work also develops spatiotemporal models for cross-region dependencies.
However, compared to univariate epidemic forecasting, evaluation of spatial epidemic forecasting remains limited and less standardized.

\paragraph{Spatiotemporal forecasting methods}

A large literature models spatiotemporal forecasting with graph structure or other spatial priors.
Representative methods include diffusion- and recurrence-based models such as DCRNN \citep{li2017diffusion} and AGCRN \citep{bai2020adaptive}; graph-convolutional architectures such as STGCN \citep{yan2018spatial}, GraphWaveNet \citep{wu2019graph}, and STNorm \citep{deng2021st}; graph-learning methods such as MTGNN \citep{wu2020connecting} and GTS \citep{shang2021discrete}; and spectral approaches such as StemGNN \citep{cao2020spectral}.
Most were developed for non-epidemic tasks, especially traffic forecasting, and sometimes broader multivariate forecasting \citep{liang2022basicts}.
Recent epidemiology-specific models, including EpiGNN \citep{xie2022epignn}, Cola-GNN \citep{deng2020cola}, EpiCola-GNN \citep{liu2023epidemiology}, and EARTH \citep{wan2025earth}, adapt similar machinery to regional infectious-disease forecasting.
They typically add epidemic-specific components, such as mechanistic structure or auxiliary regularization, and use geographic adjacency as spatial input, sometimes learning a richer network on top.
Other epidemiological models, such as CausalGNN \citep{wang2022causalgnn}, STAN \citep{gao2021stan}, STOEP \citep{ruan2026prior}, and mechanistic approaches \citep{gao2023evidence}, can use auxiliary inputs including population statistics, mobility networks, and multiple data streams.
We focus on forecasting one stream from historical data and geographic adjacency alone, since richer inputs such as mobility are not always publicly available across regions and times, e.g., SafeGraph mobility data are no longer updated after 2022, while adjacency remains popular because epidemics are expected to spread among nearby populations \citep{liu2024review}.
It remains unclear whether general spatiotemporal forecasters can be adapted to epidemics through modular changes or require more fundamental redesign.
Overall, spatiotemporal epidemic forecasting evaluations remain heterogeneous, with no widely adopted benchmark centered on operational settings epidemiologists care about, such as retraining, rolling evaluation, and outbreak-focused assessment \citep{cramer2022united}.

\section{\method{}}

\subsection{Evaluation overview}

\paragraph{Problem formulation} We consider spatiotemporal forecasting over a set of $n=|V|$ spatial units connected by a graph $G=(V,E)$ with an associated adjacency matrix $\mathbf{A}\in\mathbb{R}^{n\times n}$, where $A_{ij}$ encodes the specified relation from region $j$ to region $i$ under the chosen graph convention, and each node corresponds to a geographic region (e.g., a U.S.\ state). In this paper, the adjacencies are defined by geographic neighbors, although \method{} allows the specification of other networks, such as mobility or contact networks \citep{lyu2025combining}. At each time $t$, we observe a node-level signal $\mathbf{x}_t\in\mathbb{R}^{n}$, where $x_{t,i}$ denotes the observed value for region $i$. Given a lookback window of length $L$, the input at forecast origin $t$ is $\mathbf{X}^{(t)}
=
\begin{bmatrix}
(\mathbf{x}_{t-L+1})^\top \\
\cdots \\
(\mathbf{x}_{t})^\top
\end{bmatrix}
\in \mathbb{R}^{L\times n}$. For each forecast horizon $h$, measured in dataset-native time steps, the prediction target is $\mathbf{y}^{(t,h)} = \mathbf{x}_{t+h} \in \mathbb{R}^{n}$. The goal is to learn a forecasting model $f_{\theta_h}$ that maps historical spatiotemporal observations, and optionally graph or temporal covariates, to future node-level outcomes $\hat{\mathbf{y}}^{(t,h)} = f_{\theta_h}(\mathbf{X}^{(t)}, \mathbf{A})$. For point forecasting at horizon $h$, the base model minimizes $\mathcal{L}_{\mathrm{base}}^{(h)}(\theta_h)=\frac{1}{|\mathcal{D}_h|}\sum_{t\in\mathcal{D}_h}\ell\left(f_{\theta_h}(\mathbf{X}^{(t)}, \mathbf{A}),\mathbf{y}^{(t,h)}\right)$, where $\mathcal{D}_h$ is the set of forecast origins used for training at horizon $h$, and $\ell$ is the node-level point-forecasting loss aggregated over nodes.

\paragraph{Datasets overview} Our Delphi group collects, cleans, and maintains EpiData, a large epidemic data repository widely used for public health surveillance across multiple years and modalities \citep{farrow2015delphi}.
We prepare 11 datasets: 7 COVID-era surveillance streams from EpiData (including mixed respiratory mortality and 3 claims-based streams, DVcli, CHNGinpatient, and CHNGoutpatient), 3 COVID-19 case datasets from JHU CSSE \citep{dong2020interactive}, and 1 influenza-like illness dataset from ILINet \citep{centers2024us} (Table~\ref{tab:dataset_summary}).
Datasets were selected for long observation periods, geographic coverage, and operationally useful forecasting targets.
Although much prior forecasting work focused on influenza \citep{centers2024us}, COVID-19 expanded available data sources and modalities for measuring disease activity.
To evaluate outbreak-specific performance, we use LRTrend's outbreak annotation method \citep{lyu2025combining} to identify significantly rising intervals in each region and recompute metrics within them.
These annotations stratify evaluation only and are not model inputs.
See Appendix~\ref{app:data} for dataset characteristics and LRTrend details. We consider additional countries, diseases, and modalities in Appendix~\ref{app:extensions}(Figures~\ref{fig:supp-new-data} and~\ref{fig:supp-granularity}).

\begin{table}[hbt!]
\centering
\small
\begin{tabular}{lllll}
\toprule
Dataset & Frequency & Country & Modality & Observation period \\
\midrule
ILI2019 \citep{centers2024us} & weekly & U.S. & ILI rate & 2010--2019 \\
NCHSdeaths \citep{farrow2015delphi} & weekly & U.S. & respiratory deaths & 2020--2025 \\
CANpositivity \citep{farrow2015delphi} & daily & U.S. & test positivity & 2020--2021 \\
CHNGinpatient \citep{farrow2015delphi} & daily & U.S. & inpatient utilization & 2020--2023 \\
CHNGoutpatient \citep{farrow2015delphi} & daily & U.S. & outpatient visits & 2020--2023 \\
CPRadmissions\citep{farrow2015delphi} & daily & U.S. & hospital admissions & 2020--2023 \\
DVcli \citep{farrow2015delphi} & daily & U.S. & doctor visits & 2020--2026 \\
HHShosp \citep{farrow2015delphi} & daily & U.S. & hospitalizations & 2021--2024 \\
JHUcase \citep{dong2020interactive} & daily & U.S. & cases & 2020--2022 \\
CAcase \citep{dong2020interactive} & daily & Canada & cases & 2020--2021 \\
AUcase \citep{dong2020interactive} & daily & Australia & cases & 2020--2021 \\
\bottomrule
\end{tabular}
\caption{Summary of benchmark datasets. Periods describe the released processed files, at year granularity. Detailed signal definitions, spatial units, and zero/outlier prevalence are in Appendix~\ref{app:data}, Table~\ref{tab:zero-outlier-summary}.}
\label{tab:dataset_summary}
\end{table}

\paragraph{Evaluation metrics} We use nine point-forecasting metrics, excluding probabilistic metrics such
as WIS and CRPS because most benchmarked methods produce only point estimates, though
\method{} can incorporate them as ML methods increasingly report probabilistic forecasts.
The standard point metrics are \textbf{MSE}, \textbf{MAE}, \textbf{RMSE}, \textbf{medAE}, and \textbf{medSE}.
The filtered point metrics are \textbf{MSE filtered}, \textbf{MAE filtered}, and \textbf{RMSE filtered}which exclude zeros and outliers that may be operationally insignificant, arising from delayed or
batched reporting , holiday/weekend effects, retrospective corrections, changing case definitions
or surveillance practices, testing or ascertainment disruptions, privacy suppression of small counts,
or true low-incidence periods
We also report \textbf{win rate}, the fraction of predictions more accurate than predicting the last observed value.
Our primary evaluation uses RMSE, RMSE filtered, and win rate; all metrics appear in the Appendix.

\paragraph{Rolling evaluations}

We evaluate each method with a standardized retraining protocol that mirrors practical public-health forecasting and epidemiological benchmarks: forecasts use only data available at the forecast origin and are evaluated prospectively as new observations accrue \citep{reich2019accuracy, cramer2022pnas, cramer2022united, adiga2026idobe}.
For horizon $h$, all methods use the same input construction: a 12-observation lookback window directly predicts the target at lead $h$.
We measure $h$ in native time steps, weeks for weekly datasets and days for daily datasets, evaluating $h\in\{1,\ldots,4\}$ for weekly data and $h\in\{1,\ldots,28\}$ for daily data, consistent with short-term infectious-disease forecasting and CDC-associated evaluations \citep{cramer2022united}.
Starting from the earliest forecastable time point, models are retrained from scratch every 8 time steps using the most recent 100 observations available at that origin.
Each retraining window is split chronologically into 80\% training and 20\% validation for model fitting and selection.
The fitted model is evaluated on the next 8 forecast origins, after which the training origin advances and the process repeats.
We apply this protocol identically across methods for fair comparison under a common rolling-update setting.
This common information set and update schedule measure robustness across targets, rather than requiring every architecture to be optimal for every dataset. Appendix~\ref{app:history-seasonal} tests 200-observation fitting histories and seasonal-naive baselines (Figure~\ref{fig:supp-protocol}).
This differs from common spatiotemporal epidemiology evaluations that use a single chronological train/validation/test split, whereas forecasting hubs and outbreak benchmarks emphasize repeated prospective or expanding-window evaluation over short operational horizons \citep{deng2020cola, xie2022epignn, reich2019accuracy, cramer2022united, adiga2026idobe}.

\subsection{Epidemic priors}


We consider four model-agnostic epidemic priors, or patches, applicable to any base spatiotemporal forecaster $f_{\theta_h}$.
To our knowledge, this is the first systematic integration and evaluation of such patches for spatiotemporal epidemic forecasting.
Rather than introducing a new architecture, these patches test whether lightweight epidemiological and temporal inductive biases improve existing spatiotemporal models.
The base model takes $\mathbf{X}^{(t)}$ and outputs a point forecast $\hat{\mathbf{y}}^{(t,h)} = f_{\theta_h}(\mathbf{X}^{(t)}, \mathbf{A}) \in \mathbb{R}^{n}$ for target $\mathbf{y}^{(t,h)}\in\mathbb{R}^{n}$.
We write $\tilde{\mathbf{y}}^{(t,h)}$ for patched forecasts that directly modify predictions and $\mathbf{r}^{(t,h)}$ for auxiliary forecasts used only for regularization.

\subsubsection{Calendar covariates (TID).}
The \texttt{tid} patch adds temporal covariates: day-of-week for daily forecasts and week-of-year for weekly forecasts.
These features are common in forecasting \citep{salinas2020deepar} and can capture epidemic reporting artifacts, day-of-week effects, and seasonality \citep{rumack2023modeling, lyu2025combining}.
Let $s_{t+h}$ be the categorical time indicator for the forecast target at lead $h$.
We embed $s_{t+h}$ and pass it through a small MLP to produce a node-level additive correction $\boldsymbol{\delta}_{\phi}(s_{t+h}) \in \mathbb{R}^{n}$.
The patched forecast is $\tilde{\mathbf{y}}_{\mathrm{TID}}^{(t,h)} = \hat{\mathbf{y}}^{(t,h)} + \boldsymbol{\delta}_{\phi}(s_{t+h})$.
This model-agnostic correction applies after any node-level forecaster's forward pass.
Other parameterizations, such as multiplicative corrections, are possible; they performed similarly, so we report the additive variant.
On log-transformed targets, the additive correction becomes multiplicative on the original scale; analogous monotone adjustments apply to other variance-stabilizing transformations.


\subsubsection{Filtered loss}
The \texttt{filter} patch modifies the training loss by masking zeros and outliers that may be uninformative for operational use yet omnipresent in epidemic data \citep{rodriguez2024machine}. Such values can arise from reporting delays, batching, retrospective corrections, surveillance changes, low-incidence periods, or other data-quality artifacts rather than forecast-relevant dynamics. It excludes zero-valued targets and those outside an interquartile-range filter with threshold \(c\) (default \(c=1.5\), for a roughly normal distribution, this would correspond to about $0.7\%$ of observations excluded). 
This patch leaves the model architecture unchanged and only affects the optimization objective.

\subsubsection{Epi-aware regularization}

This patch adds an auxiliary regularization term to the base forecasting objective, introducing mechanistic epidemic structure \citep{hethcote2000mathematics, gao2021stan} without changing the base architecture.
The epidemic component is a jointly trained auxiliary branch, not a detached post-hoc model.
For each horizon $h$, the base forecaster outputs $\hat{\mathbf{y}}^{(t,h)} = f_{\theta_h}(\mathbf{X}^{(t)},\mathbf{A})$.
Let $\mathbf{r}^{(t,h)} := \mathbf{r}_{\theta_h,\phi_h}^{(t,h)}\in\mathbb{R}^{n}$ be the auxiliary structured-module output, with $\phi_h$ denoting epidemiological-branch parameters.
The patched objective is
$\mathcal{L}^{(h)}(\theta,\phi) = \frac{1}{|\mathcal{D}_h|} \sum_{t\in\mathcal{D}_h} \left[ \ell \left(\hat{\mathbf{y}}^{(t,h)}, \mathbf{y}^{(t,h)} \right) + \lambda_{\mathrm{epi}} \ell_{\mathrm{epi}} \left( \mathbf{r}^{(t,h)}, \mathbf{y}^{(t,h)} \right) \right]$,
where $\lambda_{\mathrm{epi}}\ge 0$ controls regularization strength.

We consider three variants of this patch.

\textbf{Variant 1: SIR incidence.}
In this variant, each node is associated with latent susceptible, infectious, and removed states $\mathbf{S}^{(\tau)},\mathbf{I}^{(\tau)},\mathbf{R}^{(\tau)}\in\mathbb{R}^{n}$, where $\tau$ indexes steps within the auxiliary rollout. This variant is inspired by EINNs \citep{rodriguez2023einns}. Let $\mathbf{p}\in\mathbb{R}^{n}$ denote the regional population vector. To allow spatial coupling, we define a nonnegative row-stochastic graph mixing operator $\mathbf{P}$ from $\mathbf{A}$, for example through row normalization, and compute
$\mathbf{I}_{\mathrm{mix}}^{(\tau)}=\mathbf{P}\mathbf{I}^{(\tau)}$.

Given node-level transmission and recovery rates $\boldsymbol{\beta}^{(t)},\boldsymbol{\gamma}^{(t)}\in\mathbb{R}^{n}_+$, held fixed over the auxiliary rollout for a given $(t,h)$, and time step $\Delta t$, the auxiliary incidence at rollout step $\tau$ is $\mathbf{z}^{(\tau)} = \Delta t \boldsymbol{\beta}^{(t)} \odot \frac{\mathbf{S}^{(\tau)}}{\mathbf{p}} \odot \mathbf{I}_{\mathrm{mix}}^{(\tau)}$, where $\odot$ denotes elementwise multiplication. 
The latent states are updated by a forward Euler step: $$\mathbf{S}^{(\tau+1)} = \mathbf{S}^{(\tau)} - \mathbf{z}^{(\tau)}, \mathbf{I}^{(\tau+1)} = \mathbf{I}^{(\tau)} + \mathbf{z}^{(\tau)} - \Delta t \boldsymbol{\gamma}^{(t)} \odot \mathbf{I}^{(\tau)}, \mathbf{R}^{(\tau+1)} = \mathbf{R}^{(\tau)} + \Delta t \boldsymbol{\gamma}^{(t)} \odot \mathbf{I}^{(\tau)}.$$ The rates $\boldsymbol{\beta}^{(t)}$ and $\boldsymbol{\gamma}^{(t)}$ are learned, sample-dependent quantities predicted from the same lookback window used by the base forecaster via a neural head. The auxiliary regularizer output $\mathbf{r}^{(t,h)}$ is the incidence forecast produced by this rollout at horizon $h$; under the convention that $\mathbf z^{(\tau)}$ represents incidence over the interval from \(t+\tau\) to $t+\tau+1$, we set $\mathbf r^{(t,h)}=\mathbf z^{(h-1)}$. 

\textbf{Variant 2: SIR percent.}
This variant uses the same latent rollout, but returns a population-normalized version of the auxiliary incidence forecast $\mathbf{r}^{(t,h)} = s \frac{\mathbf{z}^{(h-1)}}{\mathbf{p}}$, where division is elementwise and $s$ is a user-defined scaling constant, such as $100$ for percent or $10^5$ for per-100k units. When the auxiliary loss is computed against the correspondingly normalized target, this variant changes the regularization from a count-based penalty to a rate-based penalty and therefore is not simply absorbed by $\lambda_{\mathrm{epi}}$.

\textbf{Variant 3: NGM propagation.}
This variant replaces the nonlinear latent rollout with a structured linear propagation rule \citep{diekmann2010construction} following the epidemiology-aware deep learning formulation of \citet{liu2023epidemiology}. For each fixed forecast horizon $h$, a neural epidemiological head predicts nonnegative, sample-dependent transmission and recovery parameters $\boldsymbol{\beta}^{(t,h)}, \boldsymbol{\gamma}^{(t,h)}\in\mathbb{R}_+^n$ from the same lookback window used by the base forecaster. Using the graph-derived propagation matrix $\mathbf{P}$, one representative form is $\mathbf{K}^{(t,h)} = \operatorname{diag}(\boldsymbol{\beta}^{(t,h)}) \left(\operatorname{diag}(\boldsymbol{\gamma}^{(t,h)})-\mathbf{P}\right)^{-1}$, where $\mathbf{K}^{(t,h)}\in\mathbb{R}^{n\times n}$ is the auxiliary next-generation propagation matrix. Using column-vector convention, the auxiliary output is $\mathbf{r}^{(t,h)}=\mathbf{K}^{(t,h)}\mathbf{x}_{t}$. In our setting, this auxiliary propagation is used only as a regularizer and does not replace the base forecaster.

\subsubsection{Epi-inspired neural network (EINN).}
The \texttt{einn} patch uses EINNs as an auxiliary transfer-learning prior for the base forecaster $f_{\theta_h}$ \citep{rodriguez2023einns}. EINNs first learn latent epidemic dynamics from time alone using a PINN-style time module, then regularize a feature module, instantiated from the base forecasting architecture, to map observed histories \(\mathbf{X}^{(t)}\) into the same epidemic-state space. Thus, mechanistic structure constrains the learned representation without replacing the neural forecaster.

We denote the combined ODE-residual and gradient-matching terms by $\mathcal{L}^{(h)}_{\mathrm{dyn}}$. The EINN output map produces an auxiliary forecast $\mathbf{r}^{(t,h)}\in\mathbb{R}^{n}$ parameterized by $\psi_h$, which is trained both against the target and against the base prediction $\hat{\mathbf{y}}^{(t,h)}$. The full objective is
\begin{align*}
    \mathcal{L}^{(h)}(\theta_h,\psi_h) &= \mathcal{L}^{(h)}_{\mathrm{base}}(\theta_h) + \lambda_{\mathrm{dyn}}\mathcal{L}^{(h)}_{\mathrm{dyn}}(\psi_h) \\
    &\quad + \lambda_{\mathrm{data}} \frac{1}{|\mathcal{D}_h|} \sum_{t\in\mathcal{D}_h} \left\| \mathbf{r}^{(t,h)} - \mathbf{y}^{(t,h)} \right\|_2^2\\
    &\quad + \lambda_{\mathrm{align}} \frac{1}{|\mathcal{D}_h|} \sum_{t\in\mathcal{D}_h} \left\| \hat{\mathbf{y}}^{(t,h)} - \mathbf{r}^{(t,h)} \right\|_2^2.
\end{align*}

\section{Experimental results}

\paragraph{Benchmarked methods}

We benchmark 11 spatiotemporal methods and 3 univariate baselines.
Spatiotemporal methods include general models---\textbf{DCRNN} \citep{li2017diffusion}, \textbf{AGCRN} \citep{bai2020adaptive}, \textbf{STGCN} \citep{yan2018spatial}, \textbf{GraphWaveNet} \citep{wu2019graph}, \textbf{MTGNN} \citep{wu2020connecting}, \textbf{GTS} \citep{shang2021discrete}, \textbf{StemGNN} \citep{cao2020spectral}, and \textbf{STNorm} \citep{deng2021st}---and epidemiology-specific models---\textbf{EpiGNN} \citep{xie2022epignn}, \textbf{Cola-GNN} \citep{deng2020cola}, and \textbf{EARTH} \citep{wan2025earth}; EpiColaGNN \citep{liu2023epidemiology} is equivalent to Cola-GNN with the NGM patch.
Baselines include \textbf{DLinear} \citep{zeng2023transformers}, a linear trend/seasonality decomposition model; \textbf{ARIMA} \citep{alghamdi2019forecasting}, fit separately to each series; and a \textbf{naive} persistence baseline predicting the latest observation.
Hyperparameters are selected by grid search on the training/validation split from the first retraining window and fixed for all later rolling-origin evaluations.
The protocol permits historical observations and geographic adjacency; runs time out after 24 hours.
Patches apply to all but ARIMA and naive, which are not neural networks.
See Appendix~\ref{app:method} for implementation details.

\begin{figure}[htb!]
    \centering
    \includegraphics[width=0.8\linewidth]{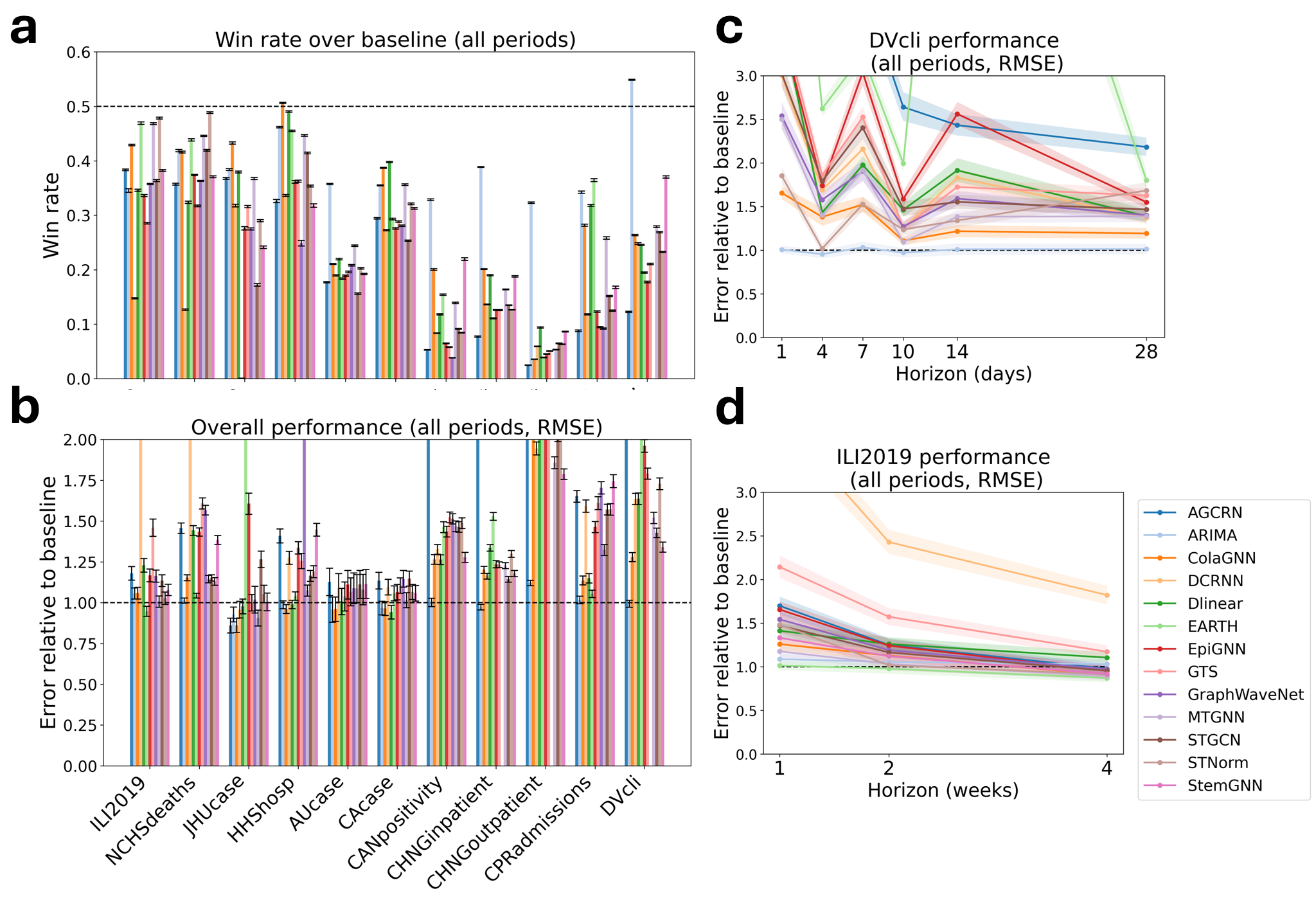}
    \caption{\textbf{Main results} \textbf{(a)} Win rate of each method versus naive baseline (higher is better), dashed line at 0.5 indicates winning 50\% of the time. \textbf{(b)} Overall RMSE relative to baseline of each method (lower is better), dashed line at 1.0 indicates equal error to baseline. \textbf{(c-d)} RMSE of each method at multiple horizons in \textbf{(c)} DVcli and \textbf{(d)} ILI2019. 95\% CI's are calculated for each plot by bootstrapping across months and meta-analyzing across horizons for \textbf{(a-b)}.
    }
    \label{fig:fig1}
\end{figure}

\paragraph{Adjacency-informed models struggle to outperform univariate methods} Figure~\ref{fig:fig1} reports results across all regions, time periods, and horizons; Appendix~\ref{app:fig1} shows similar patterns for all other metrics.
Surprisingly, every method beats the naive baseline less than 50\% of the time, often far less.
Prediction error is substantially higher than naive for almost all methods across datasets, with no clear winner over the baseline.
Even for the same disease and time period, COVID-19, the same method can perform very differently across data modalities, e.g., DVcli versus JHUCase, motivating diverse forecasting targets.
We examine horizon-specific performance in two representative datasets: smooth weekly US influenza (ILI2019) and noisy daily US COVID-19 doctor visits (DVcli).
On ILI2019, all methods underperform naive at 1 week ahead, but some slightly outperform it at 4 weeks ahead.
On DVcli, performance is much noisier, especially at 7-, 14-, and 28-day horizons, likely because weekday effects dominate the signal; no method clearly beats naive.
Overall, both univariate and adjacency-informed forecasting methods leave substantial room for improvement.

\begin{figure}[htb!]
    \centering
    \includegraphics[width=0.75\linewidth]{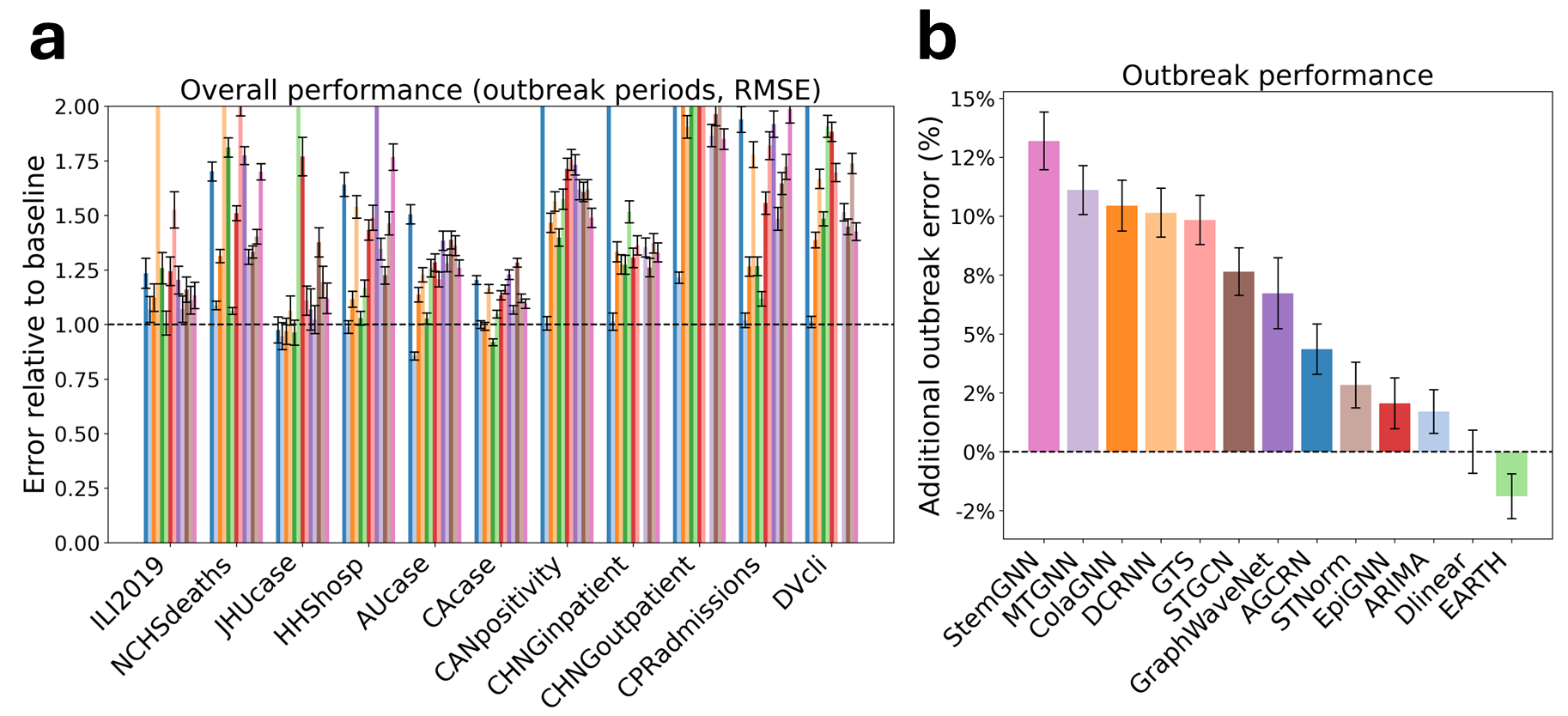}
    \caption{\textbf{Outbreak results} \textbf{(a)} Overall RMSE relative to baseline of each method (lower is better), dashed line at 1.0 indicates equal error to baseline. \textbf{(b)} Additional RMSE relative to baseline in outbreak periods versus all periods for each method. 95\% CI's are calculated for each plot by bootstrapping across months and meta-analyzing across horizons for \textbf{(a)} and datasets for \textbf{(b)}.
    }
    \label{fig:fig2}
\end{figure}

\paragraph{Adjacency-informed models struggle to predict outbreaks} We repeat the overall analysis only during outbreak periods in each region (Figure~\ref{fig:fig2}, Appendix~\ref{app:fig2}), where adjacency information should be especially useful because neighboring states may provide early warning signals.
Although predicting the last observation should intuitively fail while cases rapidly increase, almost no method outperforms naive.
All but two methods perform significantly worse than naive during outbreaks, especially adjacency-informed methods, with up to 12\% higher error.
We conclude the unique challenge of forecasting outbreaks remains difficult for all methods benchmarked.

\begin{figure}[htb!]
    \centering
    \includegraphics[width=0.8\linewidth]{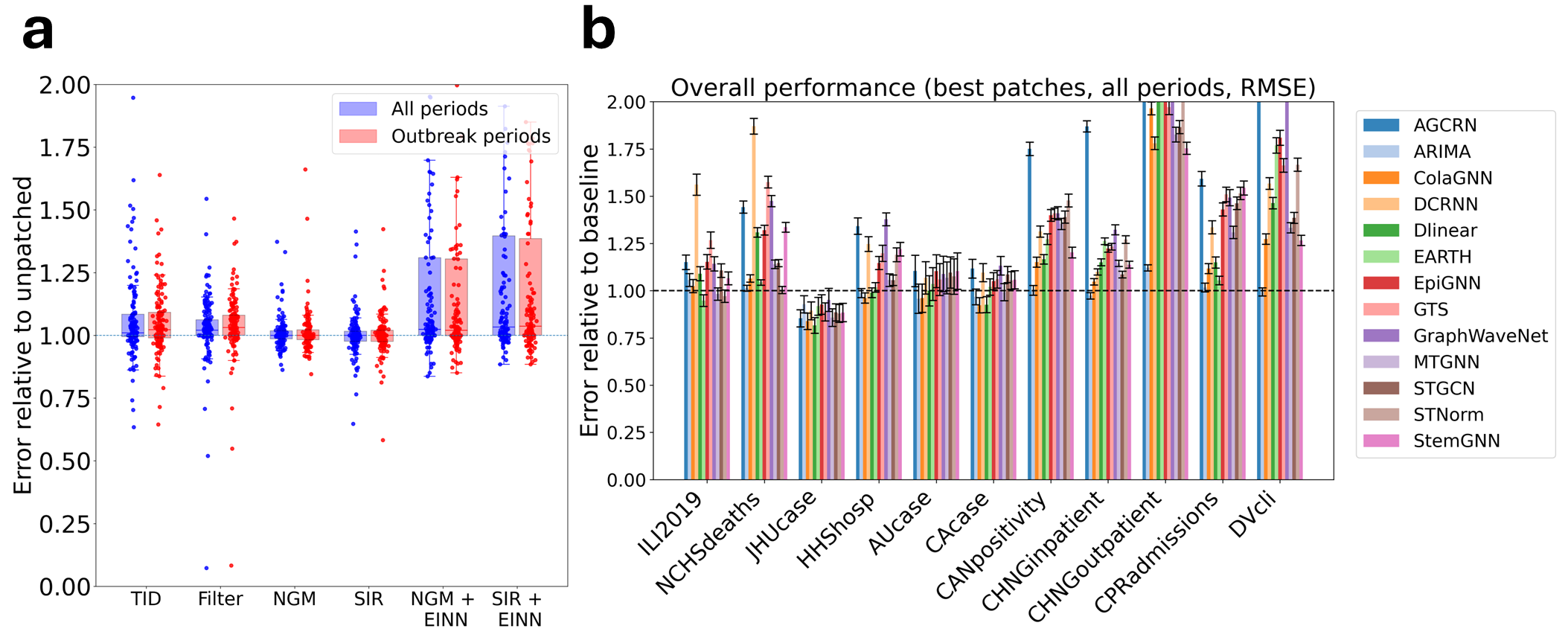}
    \caption{\textbf{Epidemic prior performance.}
    \textbf{(a)} RMSE for each method--patch--dataset combination relative to the unpatched method.
    \textbf{(b)} RMSE relative to the naive baseline for each method with its best patch configuration; lower is better, and the dashed line at 1.0 marks equal error to baseline.
    95\% CIs are computed by bootstrapping across months and meta-analyzing across horizons.
    }
    \label{fig:fig3}
\end{figure}

\paragraph{Epidemic priors selectively improve performance} We measure the improvement from all proposed patches for each method and dataset in Figure~\ref{fig:fig3} and Appendix~\ref{app:fig3}.
Calendar covariates (TID) and filtering zeros/outliers from training (Filter) can yield large method- and dataset-specific gains, lowering RMSE by $>20\%$, while NGM, SIR, and EINN often give more modest 5--15\% improvements.
However, no patch works universally, and naively adding a patch to a model more often hurts performance.
Treating patches as hyperparameters, we search for the best patch combination for each method and dataset, representing an optimistic setting where these choices are correct.
Best-patched methods substantially close the gap to naive, lowering RMSE by an average of 9\% and increasing win rates by 15\%.
Some best-patched methods now outperform naive, even in high-error settings such as JHUCase.
Still, many methods underperform naive, and adjacency-informed methods show no consistent advantage over univariate baselines.
Repeating the analysis during outbreaks gives similar conclusions and the same underperformance, even when selecting epidemic priors specifically for outbreak performance (Appendix~\ref{app:fig3}).

\begin{figure}[htb!]
    \centering
    \includegraphics[width=0.75\linewidth]{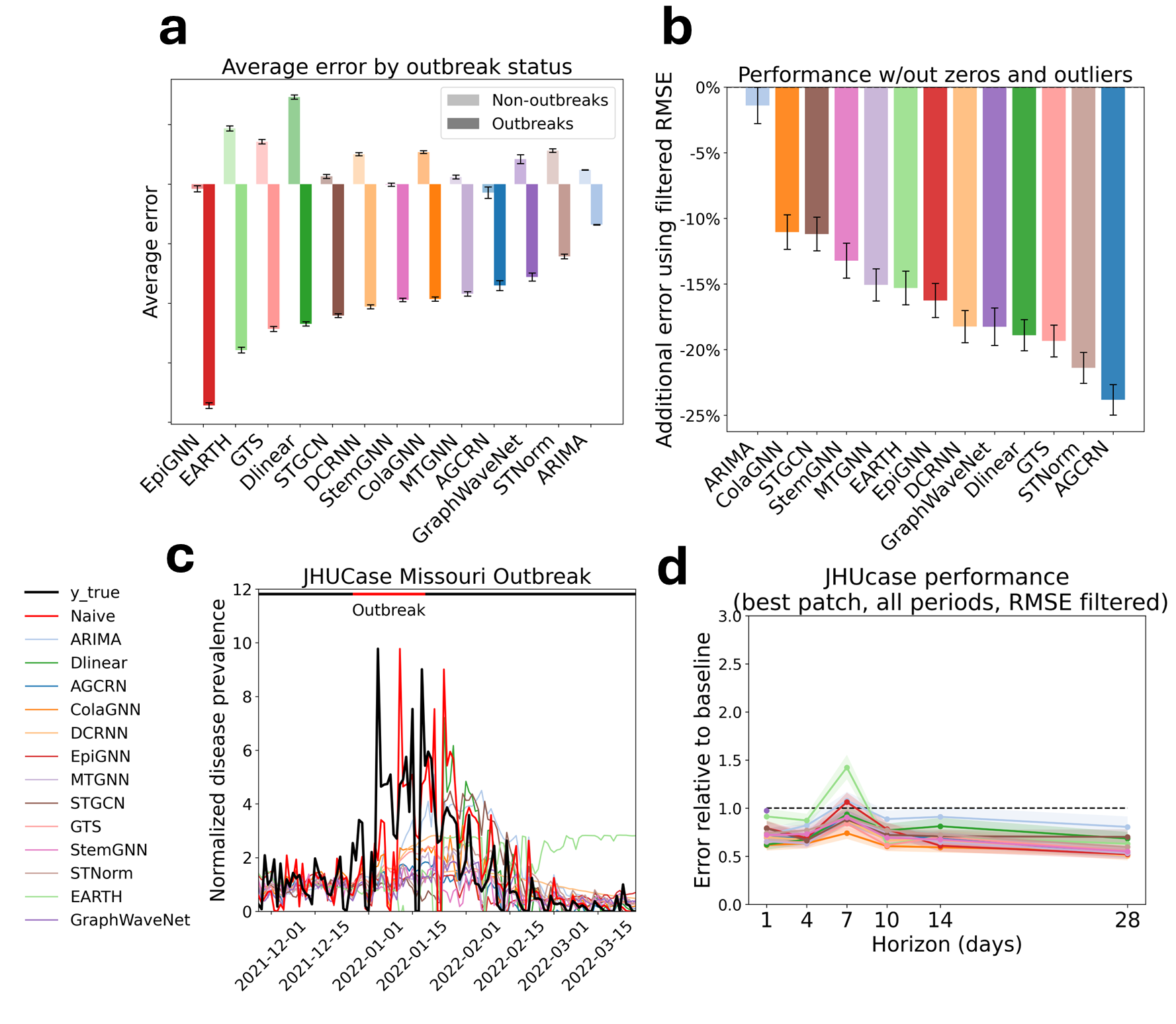}
    \caption{\textbf{Failure modes.}
    \textbf{(a)} Average signed error during outbreak and non-outbreak periods.
    \textbf{(b)} RMSE filtered versus RMSE, relative to the naive baseline.
    \textbf{(c)} Representative outbreak and method predictions at a 7-day horizon.
    \textbf{(d)} RMSE filtered for each method with its best patch at each horizon in JHUCase; lower is better.
    For all plots, 95\% CIs are computed by bootstrapping across months; for \textbf{(a--b)}, estimates are meta-analyzed across horizons and methods.
    }
    \label{fig:fig4}
\end{figure}

\paragraph{Why do models fail?} We characterize failure modes not recovered by epidemic priors.
Using each method's best-patch configuration, we identify two main causes of loss (Figure~\ref{fig:fig4}, Appendix~\ref{app:fig4}).
First, we measure average signed error in outbreak and non-outbreak periods to assess bias.
All methods strongly underpredict during outbreaks, while slightly overpredicting outside outbreaks.
A representative JHUCase example, a month-long COVID-19 outbreak around New Year's Eve in Missouri, shows several methods forecasting only small increases, while others detect the outbreak only near its end.
Second, we evaluate RMSE filtered, which excludes zeros and outliers.
Many methods improve by 10--25\% after excluding these points versus just using RMSE.
In JHUCase, the best-patch configuration on filtered points outperforms naive at nearly every horizon, though this is not true for every dataset (Appendix~\ref{app:fig4}), suggesting epidemic priors may help in some operationally relevant time points.
However, adjacency-informed methods still do not significantly outperform univariate baselines overall.

\begin{figure}[htb!]
    \centering
    \includegraphics[width=0.75\linewidth]{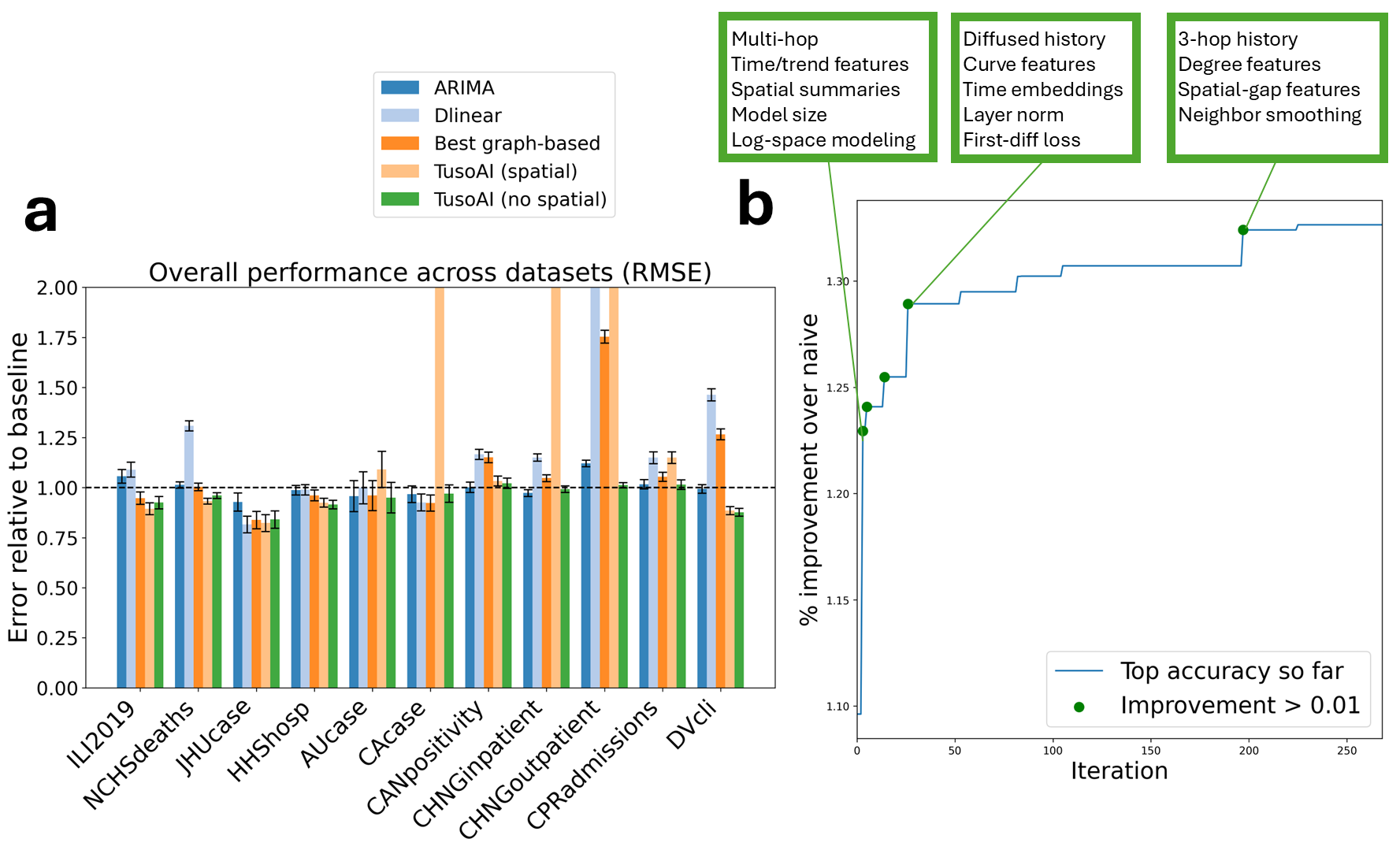}
    \caption{\textbf{Learning a new adjacency-informed model} \textbf{(a)} RMSE relative to baseline of each method with its best possible patch configuration (lower is better), dashed line at 1.0 indicates equal error to baseline. Best graph-based refers to the top spatial model in a dataset. \textbf{(b)} Optimization trajectory of TusoAI. The 3 discoveries with largest validation performance gain are annotated with a summary of the resulting code change.
    }
    \label{fig:fig5}
\end{figure}

\paragraph{Is geographic adjacency useful?} Few adjacency-informed methods match univariate baselines, and most underperform, questioning the common use of geographic proximity for epidemic forecasting.
We sought to use \method{} as a sandbox to build models which (1) outperform univariate baselines and (2) take strong advantage of adjacency information to do so.
We run the agentic method-optimization framework TusoAI \citep{turcan2025tusoai}, initialized with a simple GNN and specific instruction to leverage geographic adjacency.
TusoAI hill-climbs model code on ILI2019 and the selected architecture is evaluated across the original 11 datasets, preventing overfitting to COVID-specific properties and requiring generalizable spatiotemporal techniques.
After 24 hours, it produces a forecasting architecture with complex temporal and adjacency features, including multi-hop aggregation/smoothing and epidemiology-specific temporal features and embeddings.
The model outperforms naive on 6 datasets, including 4 where the best adjacency-informed benchmark method did not beat univariate baselines (Figure~\ref{fig:fig5}, Appendix~\ref{app:fig5}).
However, removing all spatial modeling tricks from TusoAI's model shows adjacency modeling was not critical and often reduced performance.
The run explored over 250 candidate adjacency representations, including smoothing, multi-hop aggregation, and graph-derived features; yet the graph-free ablation retained much of the improvement.
Consistent with statistical epidemiology findings, these results suggest that geographic adjacency can be a poor proxy for epidemic similarity \citep{thivierge2025does, lyu2025combining}. Supplemental geography and stratified comparisons appear in Appendices~\ref{app:extensions} and~\ref{app:strata}.
 
\section{Discussion}


We present \method{}, a comprehensive, operationally relevant benchmark for spatiotemporal epidemic forecasting.
Across methods, we identify key improvement areas, including outbreak prediction and spatial-structure integration, and diagnose common failure modes.
Several limitations suggest future work.
First, we omit probabilistic metrics such as CRPS and WIS because most benchmarked models produce point forecasts, but these can be added as probabilistic spatiotemporal models mature.
Second, we do not study intervention policies, mobility, or richer spatial networks systematically; our codebase supports multiple features and arbitrary networks, but their availability and integration vary across targets.
Third, we primarily focus on ILI and COVID-era surveillance due to the abundance of data, although our framework reaches similar conclusions in Italy and RSV. Finalized snapshots also limit conclusions about real-time revision uncertainty.
Fourth, we evaluate a similar number of methods as related epidemic forecasting work \citep{adiga2026idobe} to test whether geographic adjacency helps beyond univariate methods; stronger univariate baselines can be incorporated as forecasting methods improve.

\bibliography{sample}

\newpage
\appendix

\section{Data overview}
\label{app:data}

Our benchmark aggregates public and semi-public epidemiological surveillance signals spanning syndromic surveillance, mortality, confirmed cases, testing, hospitalization, claims-based utilization, and outpatient symptom burden. Each dataset is treated as a spatiotemporal forecasting task with a common interface across locations and forecast dates. The collection is intended to test whether forecasting methods are robust across heterogeneous epidemic data streams, rather than optimized for a single surveillance source. Below we summarize the source, target variable, and key measurement considerations for each signal. Each dataset has outbreak regions, wherein counts are quickly increasing, annotated using LRTrend \citep{lyu2025combining} run on the full time series in each region with all default parameters.

\begin{description}
    \item[\textbf{ILI2019.}] Weekly influenza-like illness (ILI) surveillance from ILINet, as reported by the CDC Influenza Division. ILI is defined syndromically as outpatient visits involving fever plus cough or sore throat; it is not laboratory-confirmed influenza and may include other respiratory pathogens with similar symptoms. We restrict this dataset to observations through the end of 2019 to avoid the major surveillance and behavioral disruptions introduced by the COVID-19 pandemic.

    \item[\textbf{NCHSdeaths.}] Weekly deaths involving pneumonia, influenza, or COVID-19 from the National Center for Health Statistics, accessed through Delphi's \texttt{deaths\_pneumonia\_or\_flu\_or\_covid\_incidence\_num} signal. This is mixed respiratory mortality, not all-cause or COVID-only mortality. The death-certificate series is indexed by occurrence week and is subject to retrospective revisions.

    \item[\textbf{JHUcase.}] Confirmed COVID-19 cases in the United States from the Johns Hopkins University Center for Systems Science and Engineering (JHU CSSE). This signal reflects reported confirmed cases and is therefore affected by testing availability, reporting practices, and changes in ascertainment over time.

    \item[\textbf{HHShosp.}] COVID-19 hospital admissions from U.S. Department of Health and Human Services (HHS) hospital reporting. The target is the number of adult and pediatric hospital admissions with confirmed COVID-19 occurring each day. HHS hospital data are part of the public health surveillance stream and capture patient impact and hospital capacity, including confirmed and suspected COVID-19 and influenza-related admissions depending on the specific source definition.

    \item[\textbf{AUcase.}] Confirmed COVID-19 cases in Australia from JHU CSSE. This task provides an international case surveillance benchmark with reporting systems and epidemic dynamics distinct from the United States.

    \item[\textbf{CAcase.}] Confirmed COVID-19 cases in Canada from JHU CSSE. As with other confirmed case signals, observations reflect reported infections and are sensitive to testing policy, eligibility, and reporting delays.

    \item[\textbf{CANpositivity.}] U.S. state-level COVID-19 test positivity from COVID Act Now (CAN), using data available through the CDC COVID-19 Integrated County View. Here CAN abbreviates COVID Act Now, not Canada. The target is the proportion of PCR specimens with a positive result. This signal captures changes in testing and transmission jointly, and may differ from case-count targets when testing volume or test-seeking behavior changes.

    \item[\textbf{CHNGinpatient.}] Inpatient COVID-19 utilization based on de-identified Change Healthcare medical claims, processed by Delphi and smoothed using a trailing 7-day average. The target is the ratio of inpatient hospitalizations associated with confirmed COVID-19 or COVID-related symptoms. Claims-based signals provide complementary coverage to public health reporting streams but may reflect insurance, billing, and healthcare-utilization patterns.

    \item[\textbf{CHNGoutpatient.}] Outpatient COVID-19 utilization based on de-identified Change Healthcare claims, processed by Delphi and smoothed using a trailing 7-day average. The target is the ratio of outpatient doctor visits associated with confirmed COVID-19 or COVID-related symptoms. This signal captures ambulatory healthcare-seeking behavior and can respond differently from case, hospitalization, or mortality targets.

    \item[\textbf{CPRadmissions.}] Confirmed COVID-19 hospital admissions from the daily Community Profile Report (CPR) produced by the Data Strategy and Execution Workgroup of the White House COVID-19 Team, smoothed using a 7-day average. CPR combines public health surveillance information on cases, deaths, testing, admissions, healthcare resources, and vaccination. Relative to HHS and claims-based hospitalization sources, CPR provides daily county-level admissions while remaining part of the public health surveillance stream; state-level aggregates are intended to align with HHS but may differ in practice.

    \item[\textbf{DVcli.}] COVID-like illness (CLI) outpatient visit estimates from Delphi's health system partners. The target is the estimated percentage of outpatient doctor visits primarily about COVID-related symptoms, smoothed in time using a Gaussian linear smoother. This syndromic signal is designed to capture symptom burden rather than laboratory-confirmed infection.
\end{description}

These datasets differ substantially in observation process, reporting lag, revision behavior, smoothing, and sensitivity to changes in testing or healthcare utilization. This heterogeneity is deliberate: a forecasting benchmark for epidemic response should evaluate performance across multiple surveillance mechanisms, including signals that are noisy, delayed, retrospectively revised, or only indirectly related to incidence.

\begin{table}[hbt!]
\centering
\footnotesize
\setlength{\tabcolsep}{4pt}
\caption{Dataset characteristics and the zero/outlier percentages reported for the supplemental analysis. The first eleven rows are the primary grid; the final three are supplemental extensions. ILI is syndromic, and NCHSdeaths measures mixed respiratory mortality. These descriptive percentages do not identify the cause of a zero or extreme value.}
\label{tab:zero-outlier-summary}
\begin{tabular}{llllrr}
\toprule
Dataset & Country & Signal / disease & Spatial unit & Zero \% & Outlier \% \\
\midrule
JHUcase & U.S. & COVID-19 cases & State/DC & 12.05 & 10.84 \\
CAcase & Canada & COVID-19 cases & Province/territory & 38.23 & 9.87 \\
AUcase & Australia & COVID-19 cases & State/territory & 52.09 & 18.67 \\
CANpositivity & U.S. & COVID-19 positivity & State/DC & 0.73 & 3.06 \\
CHNGinpatient & U.S. & COVID-19 inpatient visits & State/DC & 5.21 & 7.46 \\
CHNGoutpatient & U.S. & COVID-19 outpatient visits & State/DC & 4.50 & 6.64 \\
CPRadmissions & U.S. & COVID-19 admissions & State/DC & 0.00 & 9.96 \\
DVcli & U.S. & COVID-like illness visits & State/DC & 0.87 & 9.74 \\
HHShosp & U.S. & COVID-19 admissions & State/DC & 0.53 & 9.85 \\
NCHSdeaths & U.S. & Pneumonia/flu/COVID deaths & State/DC & 0.04 & 10.05 \\
ILI2019 & U.S. & Influenza-like illness rate & Jurisdiction$^{a}$ & 3.41 & 6.49 \\
\midrule
ITAcase & Italy & COVID-19 cases & Region & 4.52 & 12.17 \\
RSVnet & U.S. & RSV hospitalization rate & Catchment$^{b}$ & 35.71 & 8.80 \\
HRRcase & U.S. & COVID-19 cases & Referral region & 11.81 & 11.05 \\
\bottomrule
\end{tabular}
\par\smallskip
\begin{minipage}{\linewidth}\footnotesize
$^{a}$The released ILI2019 file includes 52 jurisdiction columns (states, DC, and New York City). $^{b}$Participating RSV-NET catchments in 14 states, not complete statewide populations. Percentages reproduce the supplemental dataset summary; the metric filter uses the nonzero-target IQR rule described in the main text. Low/high stratum labels in Table~\ref{tab:supp-strata} retain the supplemental grouping rather than defining a new prevalence cutoff.
\end{minipage}
\end{table}

\newpage
\section{Method implementations}
\label{app:method}
In the reported experiments, all trainable neural methods were trained for 100 epochs on machines with 32 CPU cores and NVIDIA GeForce RTX 2080 Ti GPUs, with a 24-hour wall-clock timeout. ARIMA was fit separately for each node and retraining window using its specified univariate procedure, and naive persistence has no fitted parameters. Hyperparameters were selected by grid search using the training/validation split from the first retraining window. For EARTH, the pairwise DTW matrix was precomputed and cached before rolling evaluation. Under this protocol, GraphWaveNet exceeded the wall-clock limit for all horizons on DVcli, CHNGoutpatient, and CHNGinpatient; these runs are marked as timed out.

For ease of use, the unified \texttt{run\_retrain.py} workflow also supports CPU execution. Reproducing the results requires the 100-epoch, 12-step-lookback, 100-observation-history settings and 8-step retraining schedule above. The model hyperparameters are listed below.

\begin{itemize}
    \item \textbf{DCRNN} \citep{li2017diffusion}. Diffusion recurrent model over graph random walks. Hyperparameters: `max diffusion\_step=2`, `filter\_type=dual\_random\_walk`, `num\_rnn\_layers=2`, `rnn\_units=32`, `dropout=0.1`.

    \item \textbf{AGCRN} \citep{bai2020adaptive}. Adaptive graph convolutional recurrent network with learnable node embeddings. Hyperparameters: `rnn\_units=16`, `nlayers=2`, `embed\_dim=8`, `cheb\_k=2`.

    \item \textbf{STGCN} \citep{yan2018spatial}. Spatiotemporal graph convolution blocks for joint temporal/graph modeling. Hyperparameters: `nhids=16`.

    \item \textbf{GraphWaveNet} \citep{wu2019graph}. Dilated temporal convolutions plus adaptive graph propagation. Hyperparameters: `residual\_channels=4`, `dilation\_channels=4`, `skip\_channels=32`, `end\_channels=64`, `kernel\_size=2`, `blocks=4`, `nlayers=8`.

    \item \textbf{MTGNN} \citep{wu2020connecting}. Multivariate time-series graph neural network with learned subgraph structure. Hyperparameters: `gcn\_depth=2`, `dropout=0.2`, `subgraph\_size=3`, `node\_dim=8`, `dilation\_exponential=1`, `conv\_channels=8`, `residual\_channels=4`, `skip\_channels=8`, `end\_channels=32`, `layers=3`, `propalpha=0.05`, `tanhalpha=3`.

    \item \textbf{GTS} \citep{shang2021discrete}. Graph structure learning with sequence modeling. Hyperparameters: `rnn\_units=32`, `max\_diffusion\_step=2`.

    \item \textbf{StemGNN} \citep{cao2020spectral}. Spectral-temporal model using latent correlation and temporal decomposition. Hyperparameters: `stack\_cnt=2`, `multi\_layer=4`, `dropout\_rate=0.2`, `leaky\_rate=0.2`.

    \item \textbf{STNorm} \citep{deng2021st}. Spatiotemporal normalization architecture with stacked temporal blocks. Hyperparameters: `channels=8`, `kernel\_size=2`, `blocks=8`, `layers=2`.

    \item \textbf{EpiGNN} \citep{xie2022epignn}. Epidemic-informed GNN using multi-component hidden states. Hyperparameters: `k=5`, `hidA=32`, `hidR=4`, `hidP=1`, `n\_layer=2`, `dropout=0.2`.

    \item \textbf{Cola-GNN} \citep{deng2020cola}. Correlation-aware spatiotemporal GNN for regional forecasting. Hyperparameters: `nhid=16`, `n\_layer=2`.

    \item \textbf{EARTH} \citep{wan2025earth}. DTW-informed graph forecasting model using pairwise temporal similarity. Hyperparameters: `n\_hidden=16`, `dropout=0.2`, plus required `dtw\_matrix` cache.

    \item \textbf{DLinear} \citep{zeng2023transformers}. Decomposition-linear forecasting baseline with optional future time indicators. Hyperparameters: `emb\_dim=4`, `ti\_hidden=(8,)`, `num\_timesteps\_output=1`.

    \item \textbf{ARIMA} \citep{alghamdi2019forecasting}. Univariate ARIMA baseline per node/window. Implementation uses `ARIMA(order=(1,0,0))` and forecasts to configured horizon.

    \item \textbf{Naive persistence}. Repeat-last baseline that predicts the most recent observed value for the target step.

    \item \textbf{TusoAI} \citep{turcan2025tusoai}. The exploratory search uses a 20 USD budget and 24-hour limit, Claude 4.5 Haiku, and \texttt{min\_improvement=0.01}. It starts from a simple GNN with adjacency and TID features and explicitly seeks to exploit spatial structure while improving validation performance relative to persistence across ILI2019 horizons. Over 250 candidate representations include spatial smoothing, multi-hop aggregation, and graph-derived features. The selected architecture is evaluated across the original grid. Its \emph{No Spatial} ablation removes graph operations while retaining shared temporal parameters and each node's own history. Thus graph-free does not mean independently fitted per-location parameters. This diagnostic does not establish superiority over random search, architecture search, or human tuning under matched budgets.
\end{itemize}

\newpage
\section{Supplemental geographic and disease extensions}
\label{app:extensions}
The following targeted experiments extend the benchmark's coverage and test the geographic-resolution explanation for weak adjacency gains.

\begin{figure}[htbp]
\centering
\includegraphics[width=\linewidth]{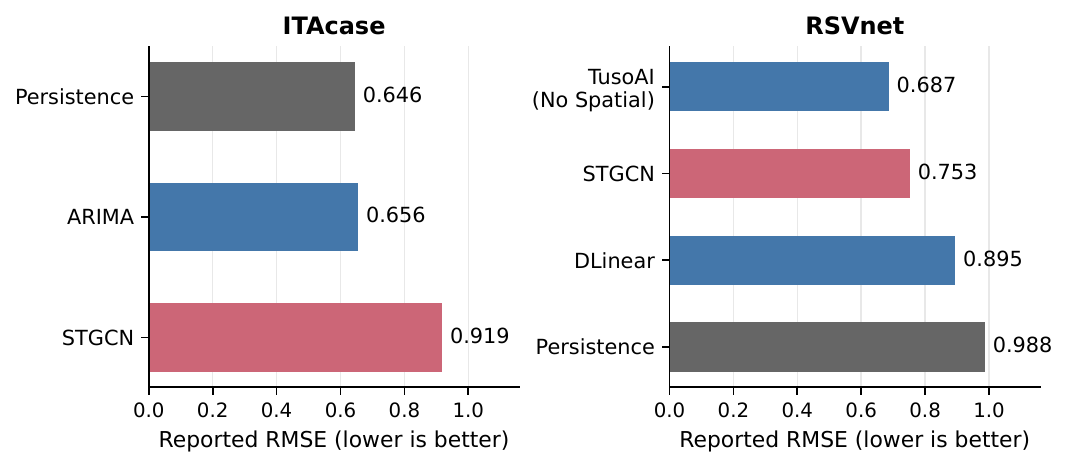}
\caption{\textbf{New geographic and disease coverage.} The reported RMSE values in Table~\ref{tab:supp-new-data} show persistence leading the Italian case comparison and graph-free TusoAI leading the RSV comparison. Colors distinguish persistence, temporal models, and the reported adjacency-informed model. Values should be compared within panels. Point estimates are shown.}
\label{fig:supp-new-data}
\end{figure}

\paragraph{Italian cases and RSV hospitalization rates.}
\textbf{ITAcase} contains daily COVID-19 cases from 20 Italian regions, February 24, 2020--March 31, 2022, using Italian Civil Protection data \citep{italianCivilProtection2026}.
\textbf{RSVnet} contains weekly RSV hospitalization rates from participating RSV-NET catchments in 14 U.S. states, October 6, 2018--December 27, 2025 \citep{cdc2026rsvnet}. The catchment definition matters: these are not whole-state hospitalization counts.
Table~\ref{tab:supp-new-data} and Figure~\ref{fig:supp-new-data} give the reported supplemental RMSE summaries. Persistence is strongest in ITAcase; the graph-free TusoAI model is strongest in RSVnet. The best adjacency-informed method is STGCN in both comparisons, but neither comparison yields a spatial-family advantage.

\begin{table}[htbp]
\centering
\small
\begin{tabular}{llrllr}
\toprule
Dataset & Method & RMSE & Dataset & Method & RMSE \\
\midrule
ITAcase & Persistence & 0.646 & RSVnet & TusoAI (No Spatial) & 0.687 \\
 & ARIMA & 0.656 & & STGCN & 0.753 \\
 & STGCN & 0.919 & & DLinear & 0.895 \\
 & & & & Persistence & 0.988 \\
\bottomrule
\end{tabular}
\caption{Reported supplemental comparisons on the two new disease/geographic streams. Values are RMSE summaries, with lower values better within each dataset. The methods listed are the reported reference methods and selected leaders. These summaries are separate from the matched 7-day analysis in Table~\ref{tab:supp-granularity}.}
\label{tab:supp-new-data}
\end{table}

\begin{figure}[htbp]
\centering
\includegraphics[width=0.94\linewidth]{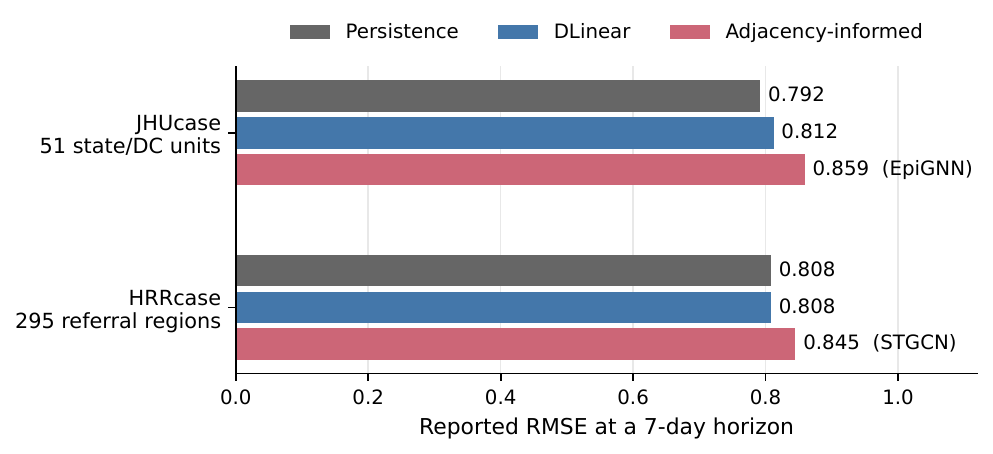}
\caption{\textbf{Finer geography alone does not establish an adjacency benefit.} At the matched 7-day horizon, the reported spatial model remains above persistence at both state and HRR scales; DLinear ties persistence at HRR scale to the reported precision. The model supplying the spatial value differs between scales and is named in the plot. The chart visualizes Table~\ref{tab:supp-granularity}.}
\label{fig:supp-granularity}
\end{figure}

\paragraph{Finer geographic resolution.}
\textbf{HRRcase} comprises daily U.S. COVID-19 cases over 295 Hospital Referral Regions (HRRs), with HRR-level adjacency. HRRs are health-care markets based on referral patterns and may cross state boundaries \citep{dartmouth2021covid}. At a matched 7-day forecast horizon, Table~\ref{tab:supp-granularity} compares the reported state-level JHUcase and HRR-level results. The best adjacency-informed method does not beat persistence at either resolution (Figure~\ref{fig:supp-granularity}). This result shows that finer resolution alone does not produce a spatial advantage in these comparisons.

\begin{table}[htbp]
\centering
\small
\begin{tabular}{llrrrr}
\toprule
Scale & Dataset & Regions & Persistence & DLinear & Adjacency-informed \\
\midrule
State & JHUcase & 51 & 0.792 & 0.812 & 0.859 (EpiGNN) \\
HRR & HRRcase & 295 & 0.808 & 0.808 & 0.845 (STGCN) \\
\bottomrule
\end{tabular}
\caption{Reported RMSE at a matched 7-day horizon. DLinear is the reported fitted temporal comparator; the final column identifies the reported leading adjacency-informed model at each scale. Persistence remains an explicit separate comparator.}
\label{tab:supp-granularity}
\end{table}

\clearpage
\section{Fitting-history and seasonal-baseline checks}
\label{app:history-seasonal}
\paragraph{Doubling the fitting history.}
We reran five representative datasets with 200 rather than 100 observations in the fitting history while retaining the 12-observation input window, forecast horizons, retraining interval, and hyperparameters. Table~\ref{tab:supp-history} gives the best overall method reported in each of these comparisons; its identity is unchanged. Scores improve on two datasets, remain unchanged on two, and worsen on RSVnet. In particular, STGCN wins the ILI2019 comparison, illustrating why our broader conclusion is an absence of \emph{consistent} spatial advantage, not that a spatial model can never win. 

\begin{table}[htbp]
\centering
\small
\begin{tabular}{llrr}
\toprule
Dataset & Best overall method at both histories & History 100 & History 200 \\
\midrule
JHUcase & DLinear & 0.914 & 0.818 \\
ILI2019 & STGCN & 0.824 & 0.811 \\
HHShosp & ARIMA & 0.497 & 0.497 \\
ITAcase & Persistence & 0.646 & 0.646 \\
RSVnet & TusoAI (No Spatial) & 0.687 & 0.711 \\
\bottomrule
\end{tabular}
\caption{Reported RMSE in the 100-versus-200-observation fitting-history experiment. Lower is better within each dataset. The winners are those reported for this experiment; its summaries should not be substituted for the separate matched 7-day scores in Table~\ref{tab:supp-granularity}.}
\label{tab:supp-history}
\end{table}

\begin{figure}[htbp]
\centering
\includegraphics[width=\linewidth]{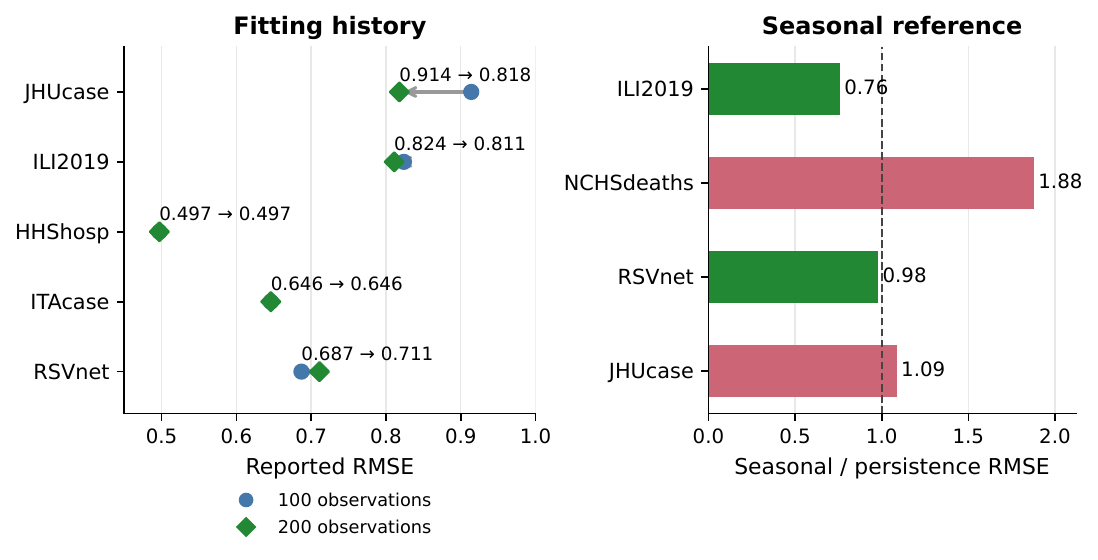}
\caption{\textbf{Protocol checks give conditional, not universal, improvements.} \textbf{Left:} RMSE with 100 and 200 fitting observations for the unchanged winning method in each dataset (Table~\ref{tab:supp-history}); arrows show the direction of change. \textbf{Right:} seasonal naive divided by persistence RMSE from Table~\ref{tab:supp-seasonal}; values below one favor the seasonal reference. Neither experiment changes the original grid.}
\label{fig:supp-protocol}
\end{figure}

\paragraph{Seasonal-naive baseline.}
A standard seasonal-naive reference repeats the most recent observed value from the same season \citep{hyndman2021fpp3}. With origin $t$, lead $h$, and seasonal period $m$, its conventional definition is
\[
\widehat{x}_{t+h}=x_{t+h-m\lceil h/m\rceil},\qquad m=7\ \text{for daily data},\quad m=52\ \text{for weekly data}.
\]
The referenced timestamp is at or before $t$, and therefore strictly precedes the target. This rule uses accessible seasonal history rather than restricting the baseline to the neural models' 12-step input window; an origin without the required prior observation cannot receive a seasonal forecast. Seasonal and persistence scores must be compared on common eligible origins. A 52-week period is an operational lag convention, rather than an exact calendar-year match in every year.
Table~\ref{tab:supp-seasonal} reports the supplemental comparison. Seasonal naive is competitive on ILI2019 and slightly improves the RSVnet score, but is worse than persistence on NCHSdeaths and JHUcase (Figure~\ref{fig:supp-protocol}).

\begin{table}[htbp]
\centering
\small
\begin{tabular}{llrr}
\toprule
Dataset & Seasonal period & Persistence & Seasonal naive \\
\midrule
ILI2019 & 52 weeks & 1.08 & 0.82 \\
NCHSdeaths & 52 weeks & 0.58 & 1.09 \\
RSVnet & 52 weeks & 0.99 & 0.97 \\
JHUcase & 7 days & 1.03 & 1.12 \\
\bottomrule
\end{tabular}
\caption{Reported RMSE summaries for the seasonal-naive comparison. These summaries are presented separately from the matched 7-day experiment: for a daily series at lead exactly 7 with $m=7$, the conventional seasonal rule above equals persistence.}
\label{tab:supp-seasonal}
\end{table}

\clearpage
\section{Descriptive stratified comparisons}
\label{app:strata}
Table~\ref{tab:supp-strata} retains the supplemental comparison of reported temporal and spatial model scores normalized by persistence, averaged across datasets in each group. The univariate score is lower in every listed stratum.

We label the twelve-stream group \emph{COVID-era surveillance}, preserving its reported aggregate while explicitly including mixed pneumonia/influenza/COVID mortality (NCHSdeaths), COVID-like illness, cases, positivity, and utilization streams. 

\begin{table}[htbp]
\centering
\small
\setlength{\tabcolsep}{4pt}
\begin{tabular}{llrrr}
\toprule
Grouping & Stratum & Datasets & Univariate score & Spatial score \\
\midrule
Disease / signal & COVID-era surveillance$^{a}$ & 12 & 1.159 & 1.817 \\
 & Influenza-like illness & 1 & 0.759 & 0.871 \\
 & RSV & 1 & 0.731 & 0.802 \\
\midrule
Modality & Cases/positivity & 6 & 1.177 & 1.675 \\
 & Visits/admissions/hospitalizations & 6 & 1.154 & 1.794 \\
 & Deaths/rates/other & 2 & 0.915 & 1.336 \\
\midrule
Frequency & Daily & 11 & 1.186 & 1.819 \\
 & Weekly & 3 & 0.922 & 1.158 \\
\midrule
Sparsity / zeros & Low (reported grouping) & 11 & 1.163 & 1.838 \\
 & High (reported grouping) & 3 & 1.005 & 1.086 \\
\midrule
Outliers & Low (reported grouping) & 7 & 1.123 & 1.761 \\
 & High (reported grouping) & 7 & 1.136 & 1.593 \\
\bottomrule
\end{tabular}
\caption{Reported supplemental stratum summaries. Scores below one improve on persistence under the reported normalization; the columns describe univariate and spatial model summaries. Low/high labels retain the reported partition.}
\label{tab:supp-strata}
\par\smallskip
\begin{minipage}{\linewidth}\footnotesize
$^{a}$All streams in Table~\ref{tab:zero-outlier-summary} except ILI2019 and RSVnet. 
\end{minipage}
\end{table}

\newpage
\section{Supporting information for Figure 1}
\label{app:fig1}

\begin{figure}[htb!]
    \centering
    \includegraphics[width=1.0\linewidth]{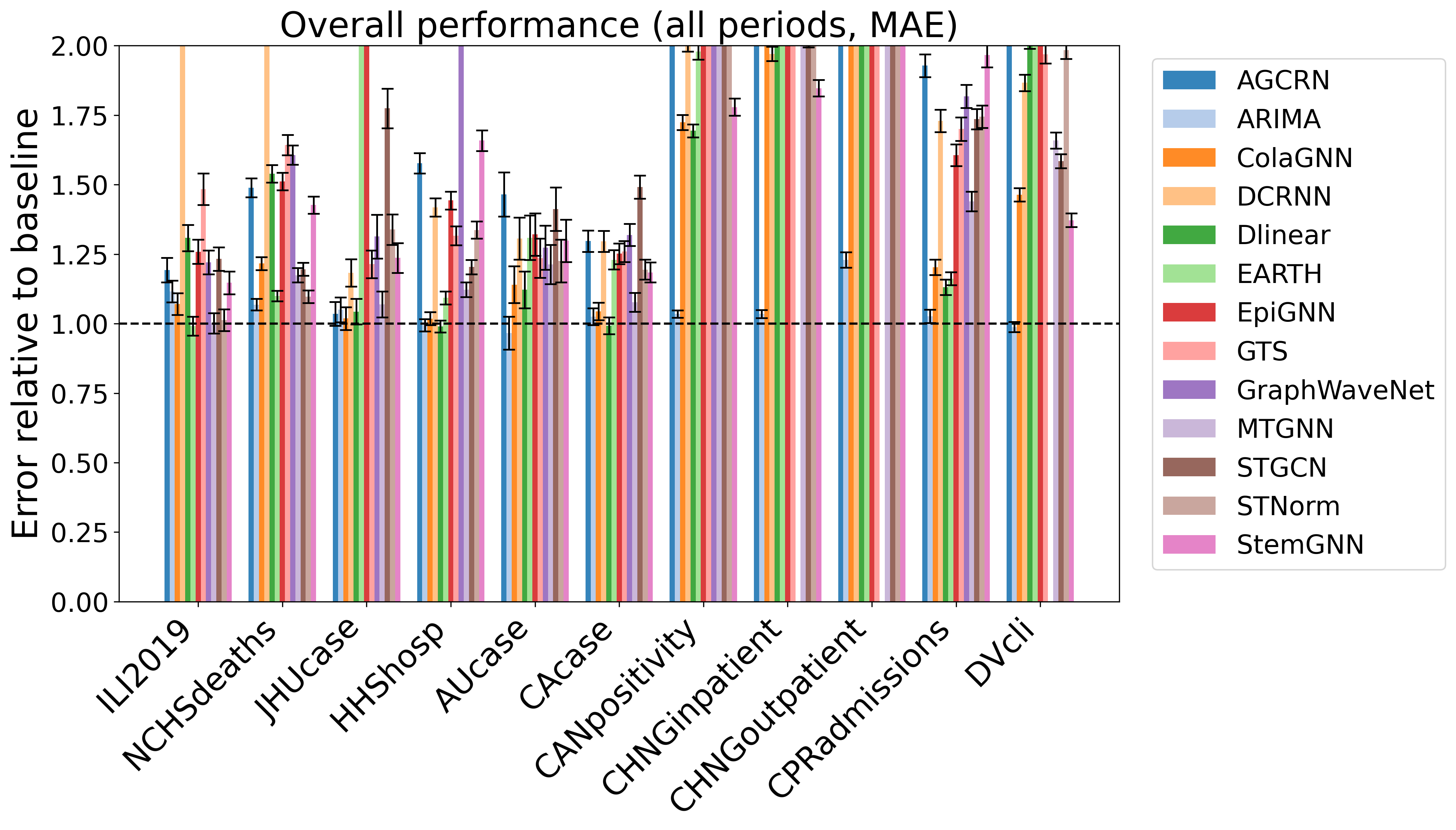}
    \caption{ Error of each method. 95\% CI's are calculated for each plot by bootstrapping across months and meta analyzing across horizons.
    }
\end{figure}

\begin{figure}[htb!]
    \centering
    \includegraphics[width=1.0\linewidth]{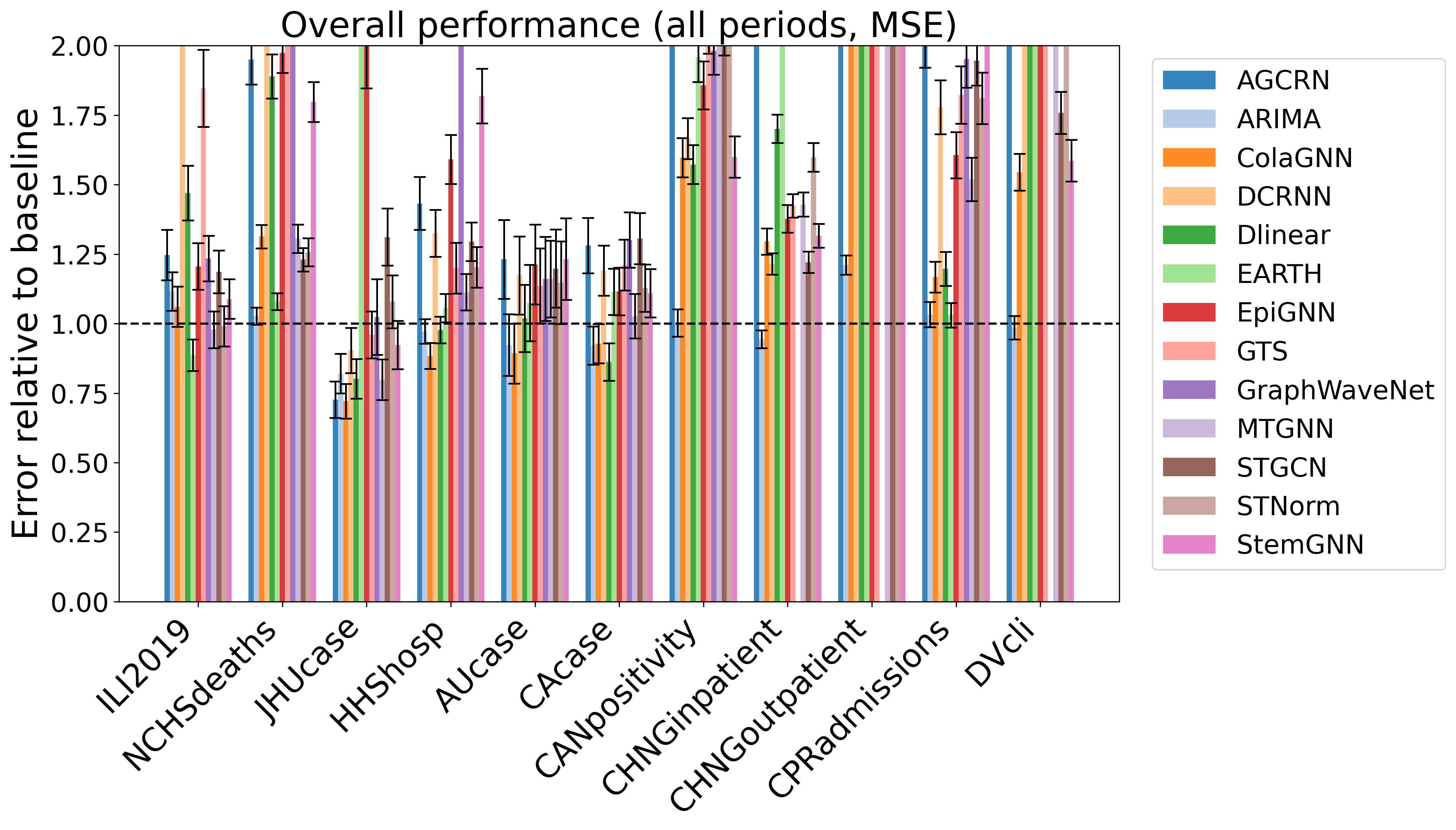}
    \caption{ Error of each method. 95\% CI's are calculated for each plot by bootstrapping across months and meta analyzing across horizons.
    }
\end{figure}
\newpage

\begin{figure}[htb!]
    \centering
    \includegraphics[width=1.0\linewidth]{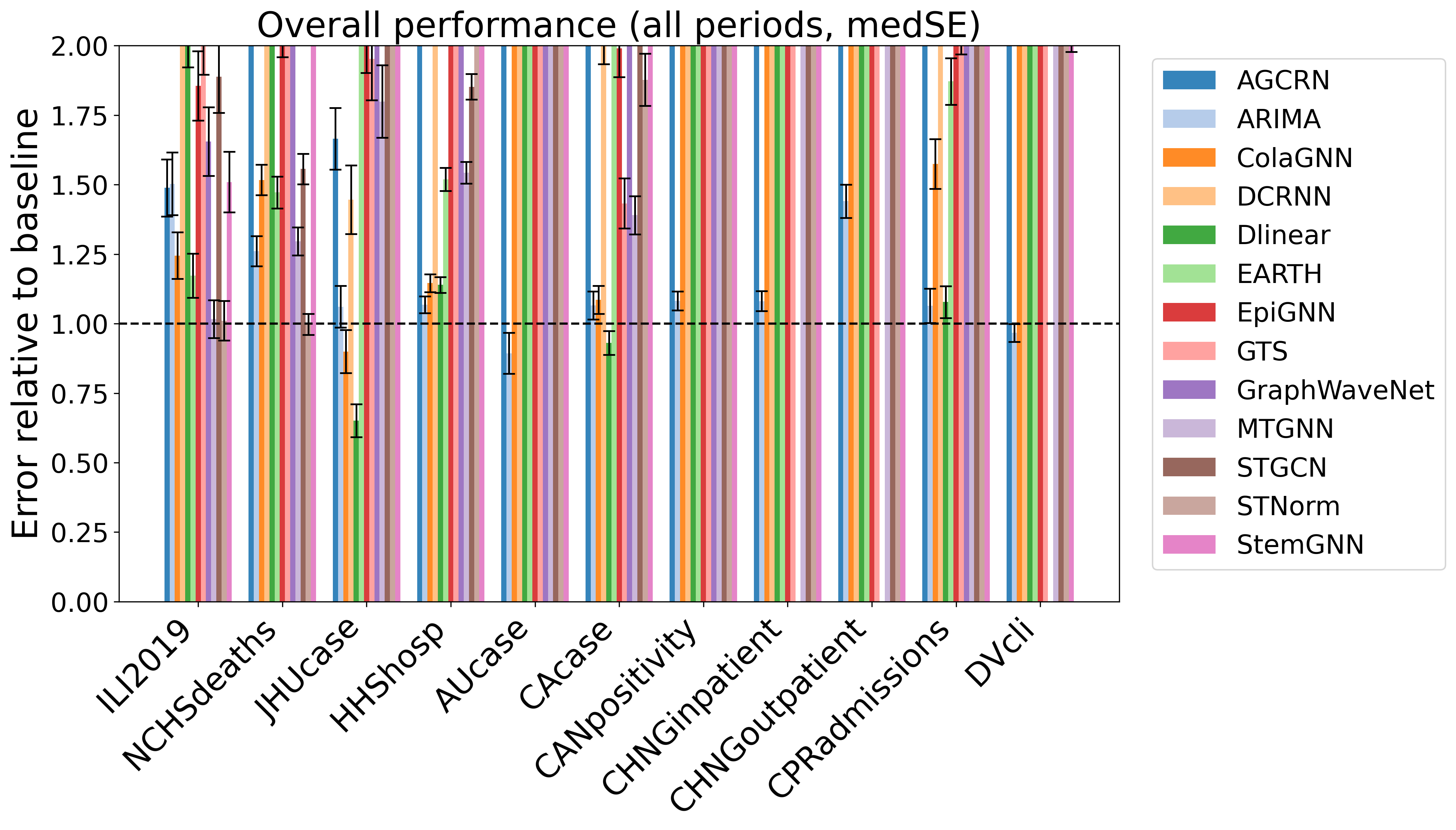}
    \caption{ Error of each method. 95\% CI's are calculated for each plot by bootstrapping across months and meta analyzing across horizons.
    }
\end{figure}

\begin{figure}[htb!]
    \centering
    \includegraphics[width=1.0\linewidth]{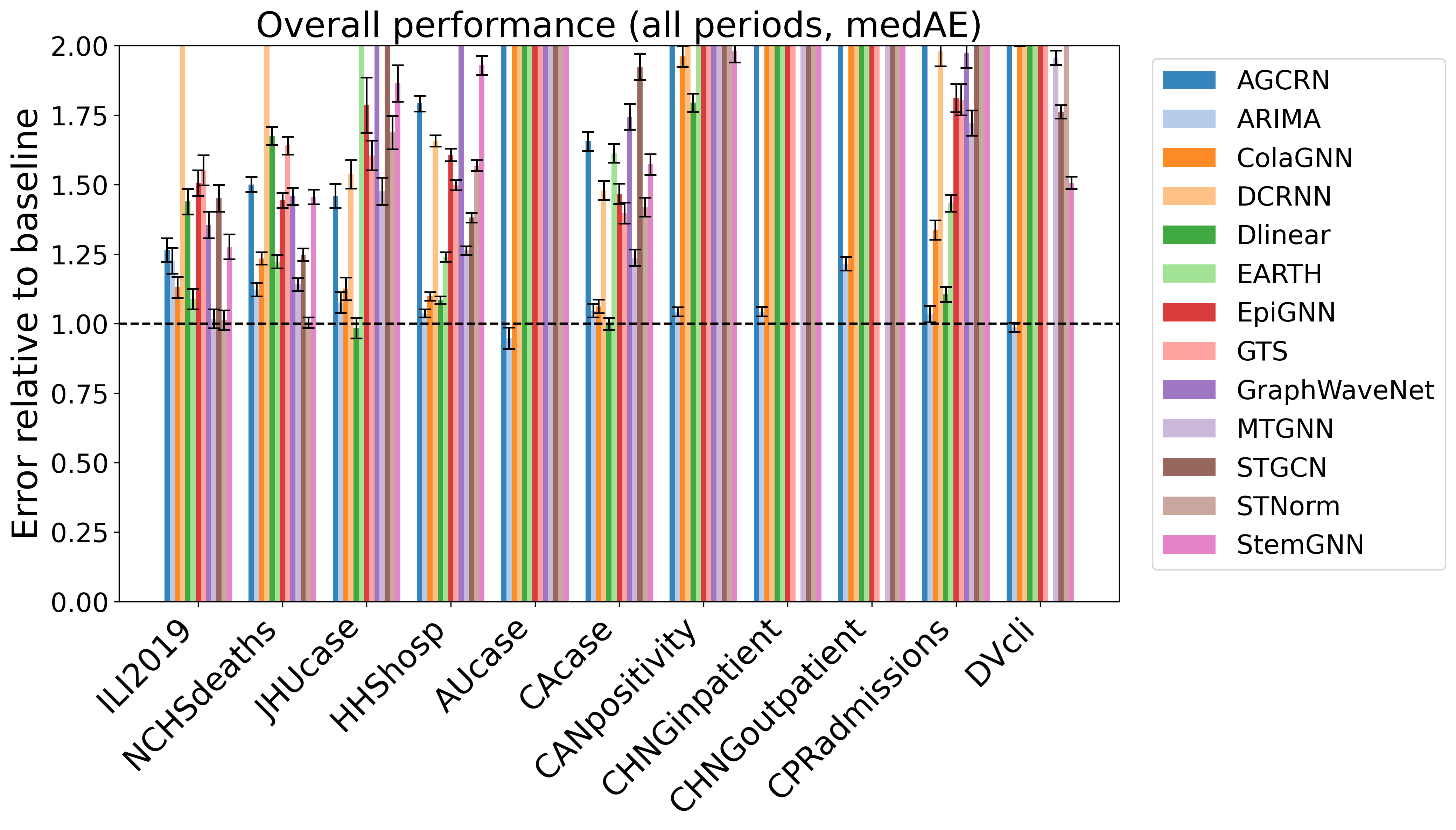}
    \caption{ Error of each method. 95\% CI's are calculated for each plot by bootstrapping across months and meta analyzing across horizons.
    }
\end{figure}
\newpage

\begin{figure}[htb!]
    \centering
    \includegraphics[width=1.0\linewidth]{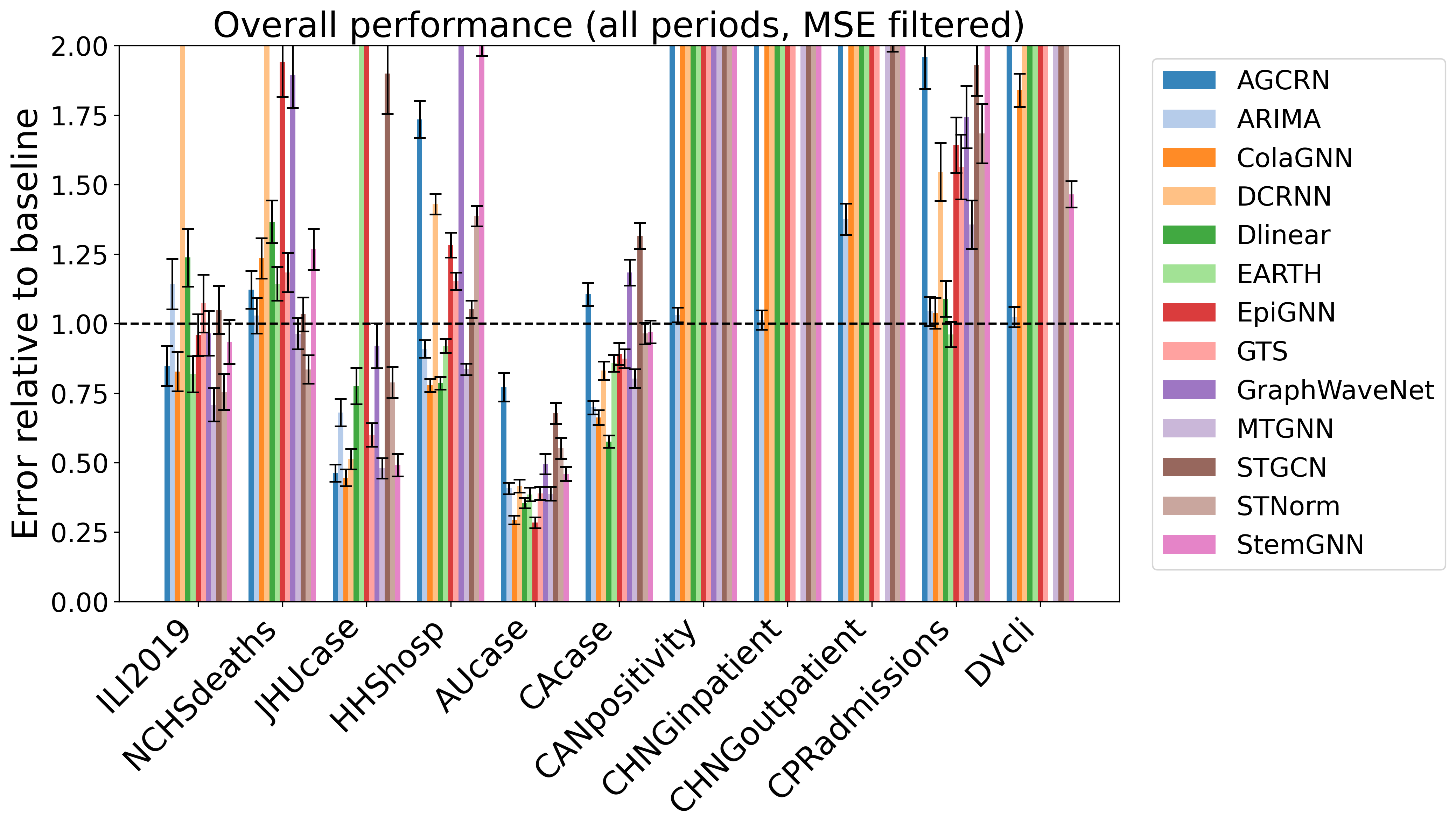}
    \caption{ Error of each method. 95\% CI's are calculated for each plot by bootstrapping across months and meta analyzing across horizons.
    }
\end{figure}
\begin{figure}[htb!]
    \centering
    \includegraphics[width=1.0\linewidth]{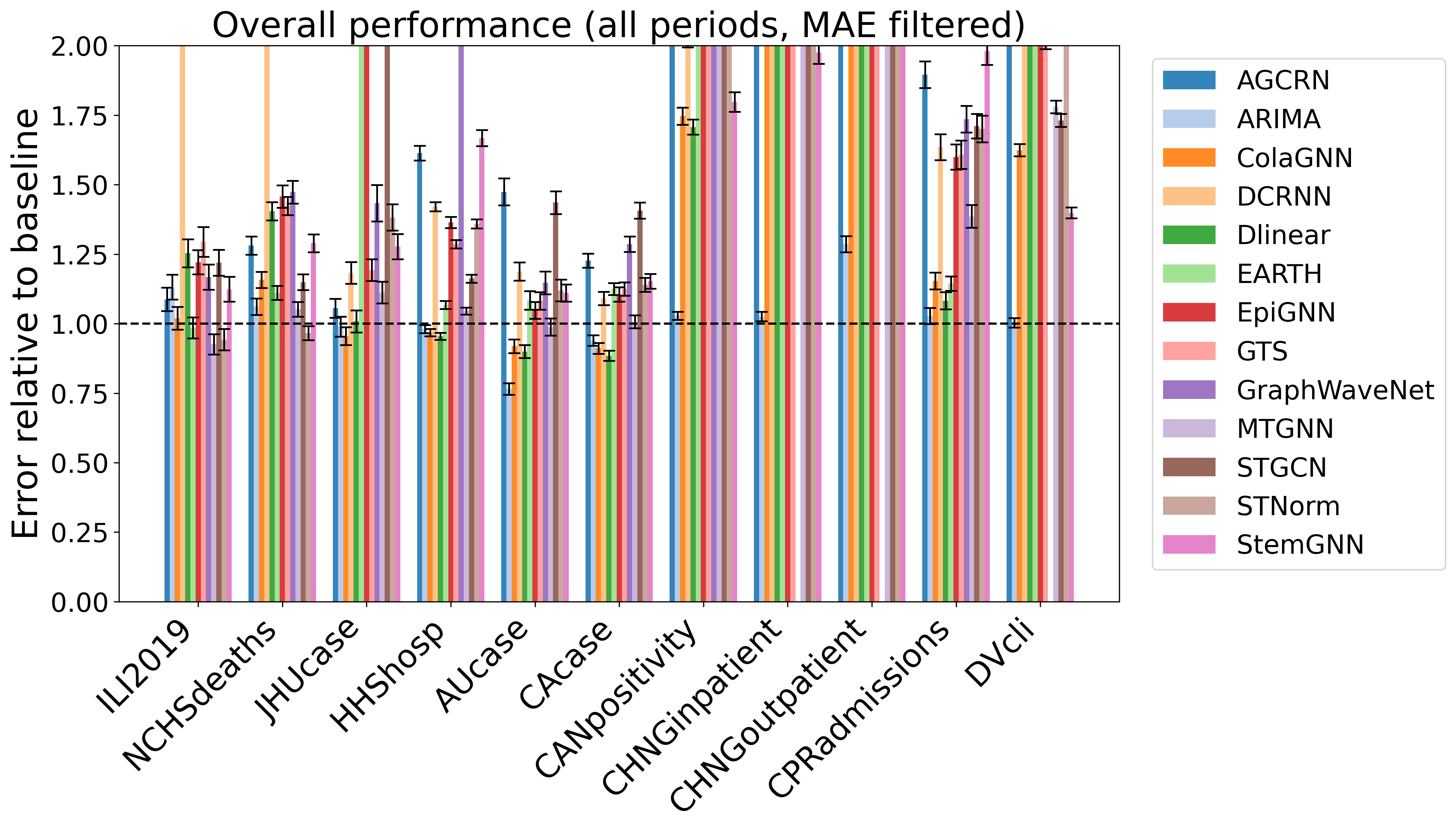}
    \caption{ Error of each method. 95\% CI's are calculated for each plot by bootstrapping across months and meta analyzing across horizons.
    }
\end{figure}
\newpage

\begin{figure}[htb!]
    \centering
    \includegraphics[width=1.0\linewidth]{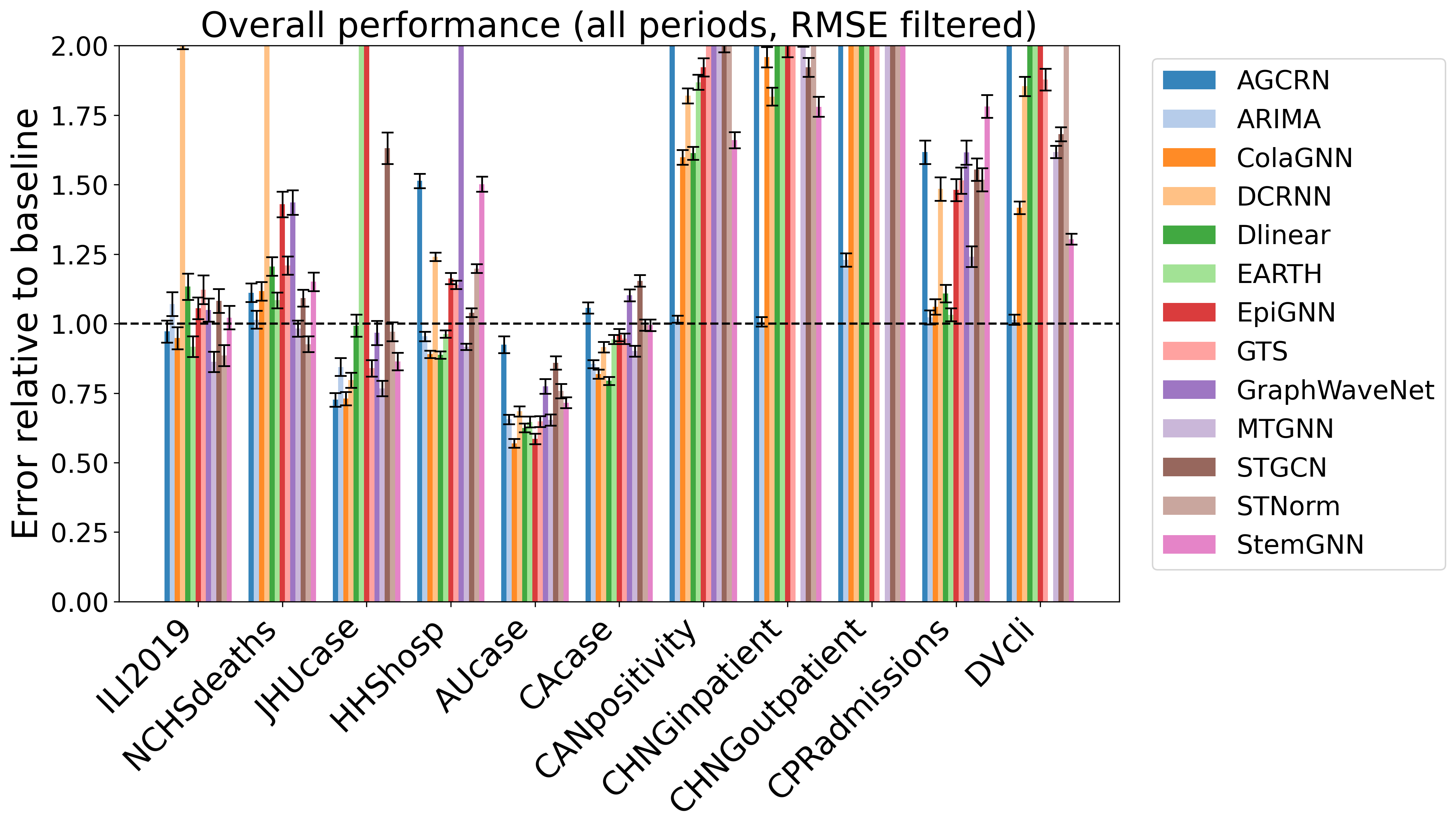}
    \caption{ Error of each method. 95\% CI's are calculated for each plot by bootstrapping across months and meta analyzing across horizons.
    }
\end{figure}

\newpage
\begin{figure}[htb!]
    \centering
    \includegraphics[width=1.0\linewidth]{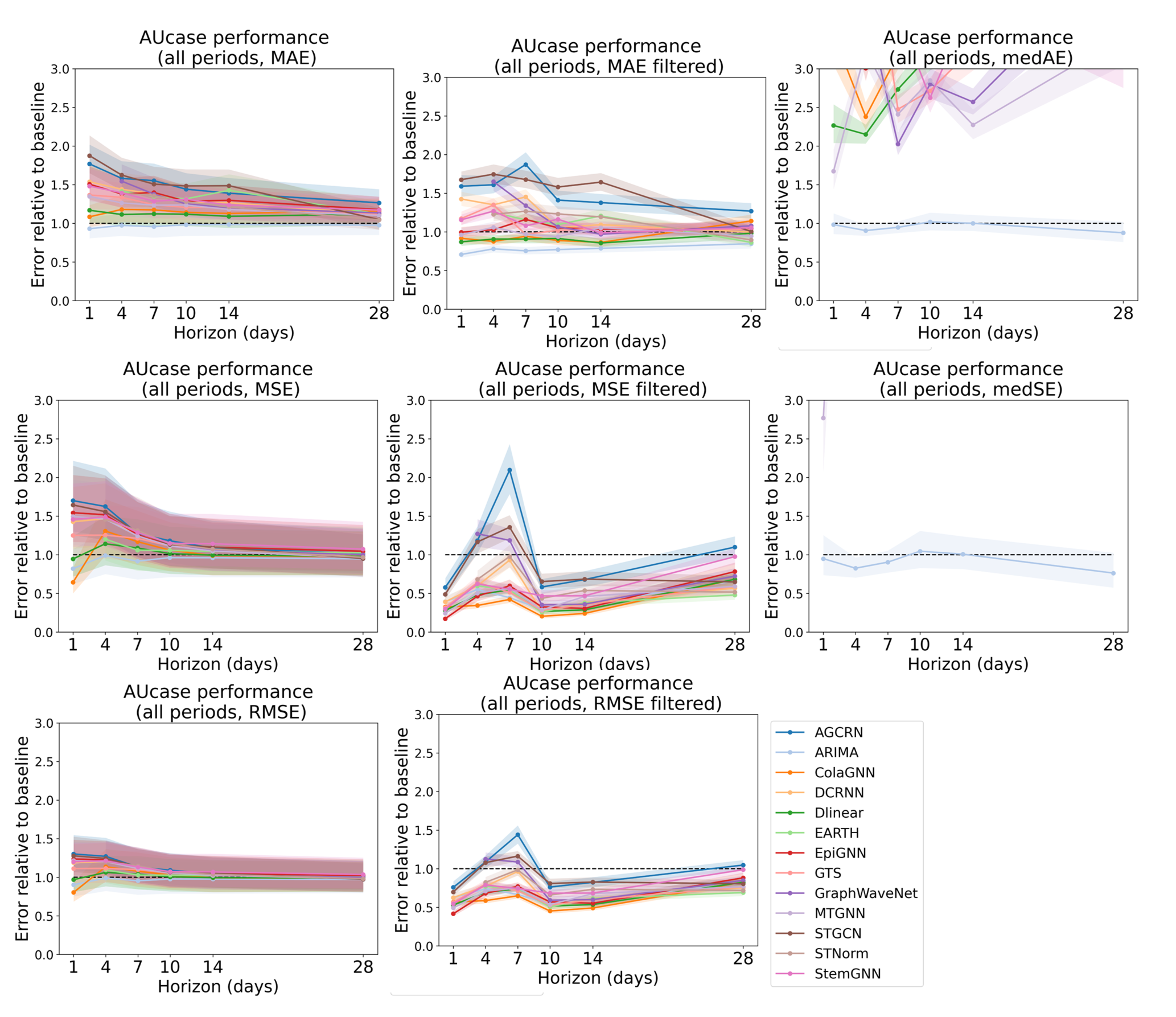}
    \caption{ Error of each method at multiple horizons in AUcase by each metric. 95\% CI's are calculated for each plot by bootstrapping across months.
    }
\end{figure}
\newpage

\begin{figure}[htb!]
    \centering
    \includegraphics[width=1.0\linewidth]{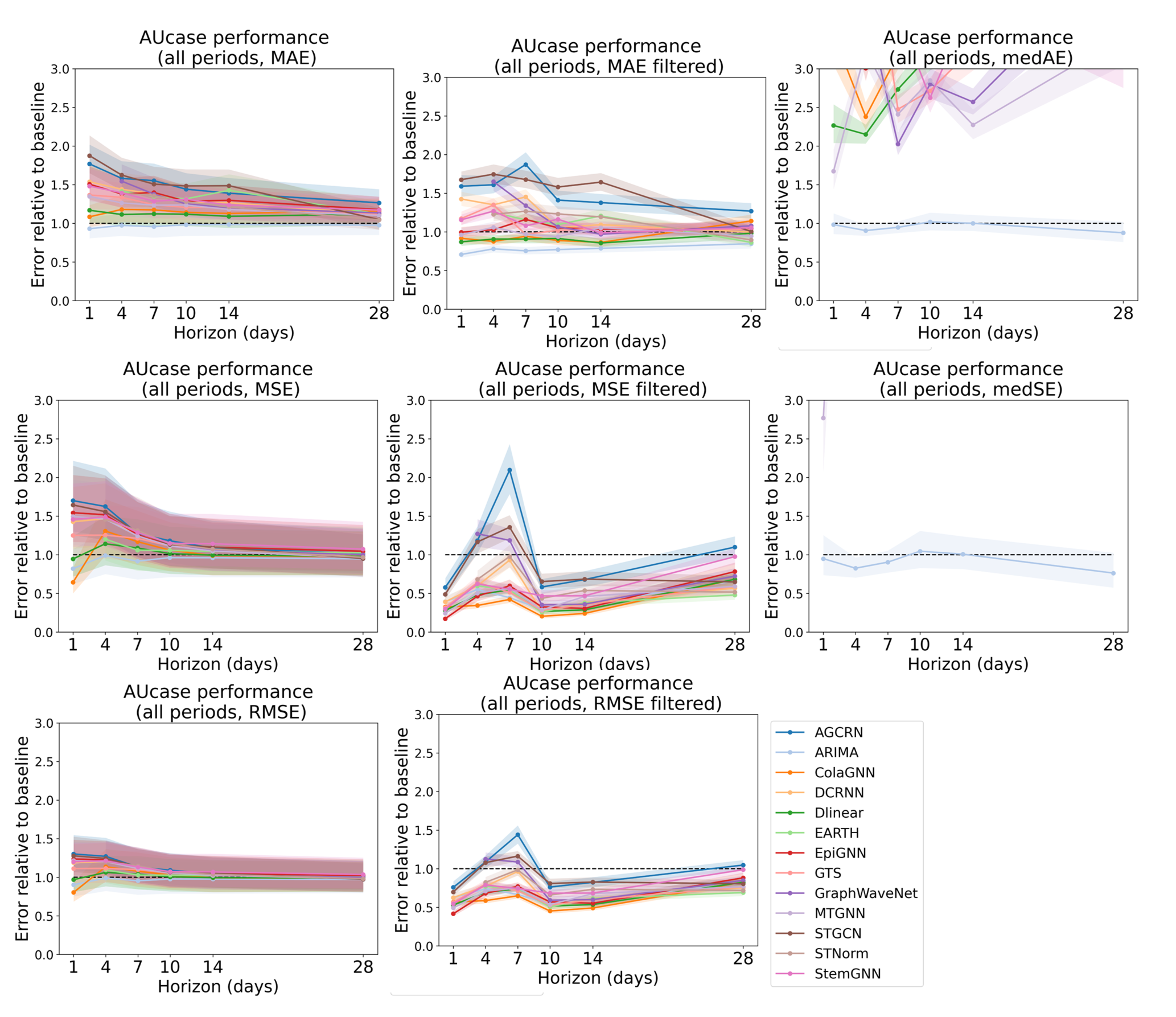}
    \caption{ Error of each method at multiple horizons in AUcase by each metric. 95\% CI's are calculated for each plot by bootstrapping across months.
    }
\end{figure}
\newpage

\begin{figure}[htb!]
    \centering
    \includegraphics[width=1.0\linewidth]{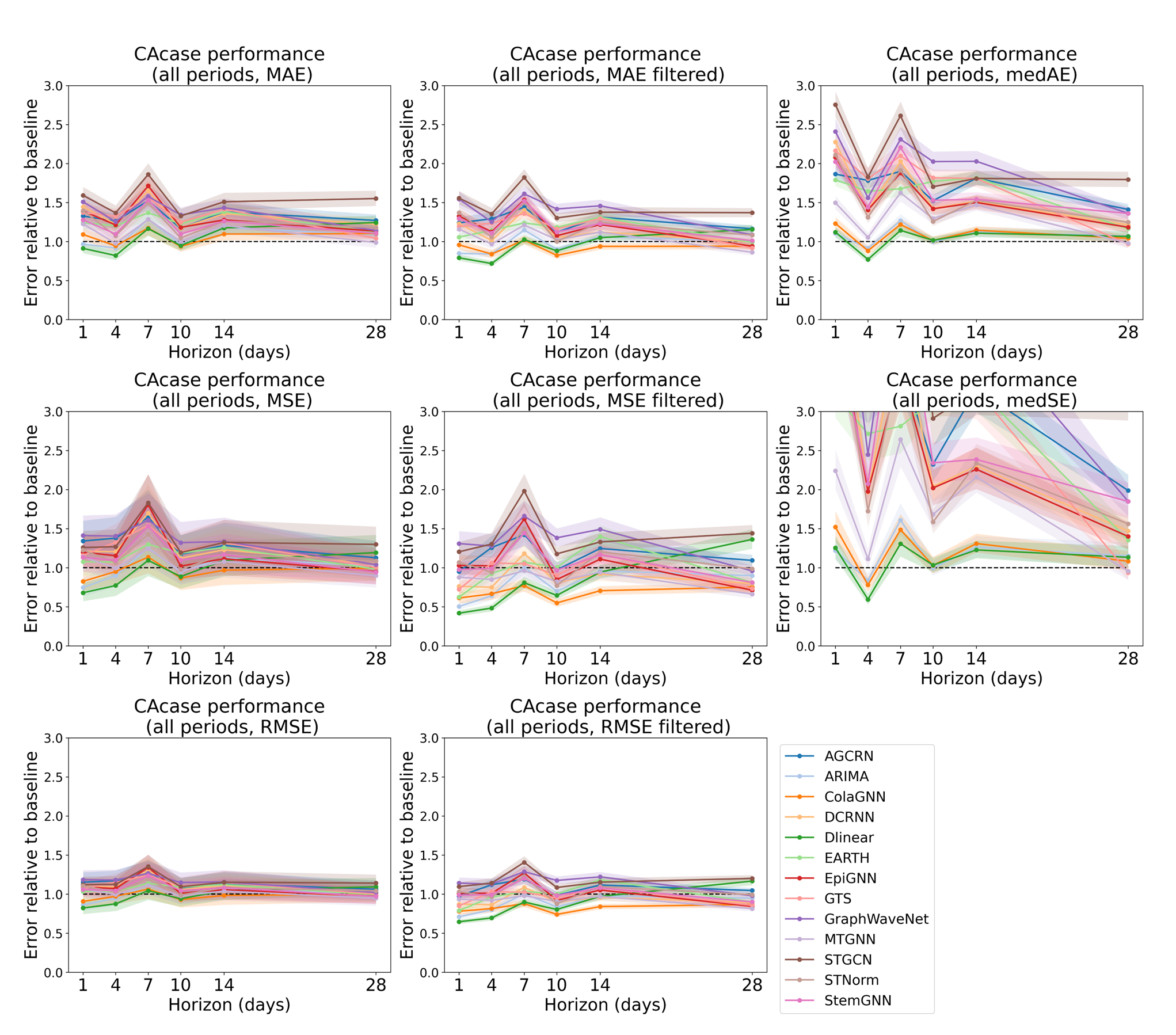}
    \caption{ Error of each method at multiple horizons in CAcase by each metric. 95\% CI's are calculated for each plot by bootstrapping across months.
    }
\end{figure}
\newpage

\begin{figure}[htb!]
    \centering
    \includegraphics[width=1.0\linewidth]{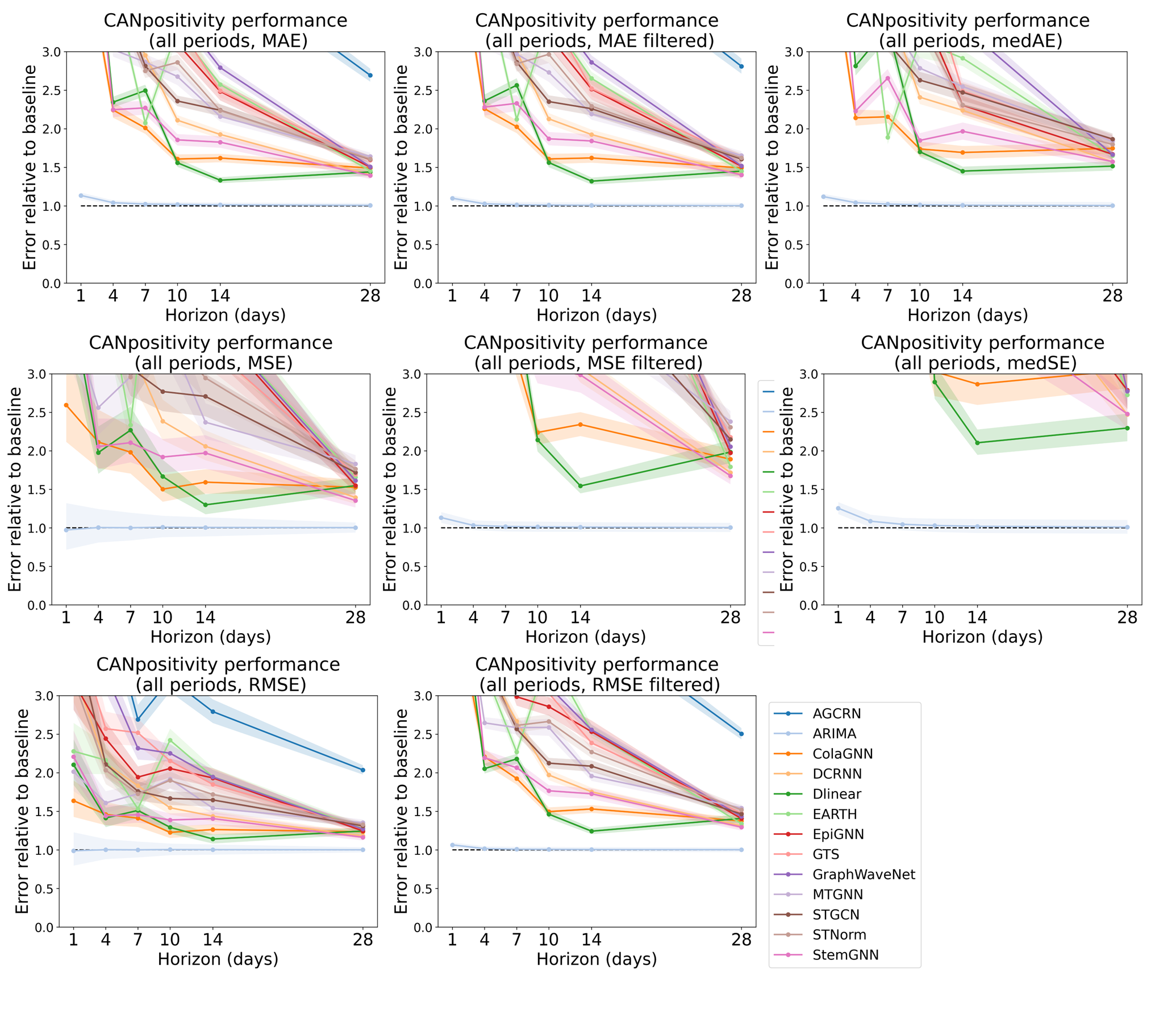}
    \caption{ Error of each method at multiple horizons in CANPositivity by each metric. 95\% CI's are calculated for each plot by bootstrapping across months.
    }
\end{figure}
\newpage

\begin{figure}[htb!]
    \centering
    \includegraphics[width=1.0\linewidth]{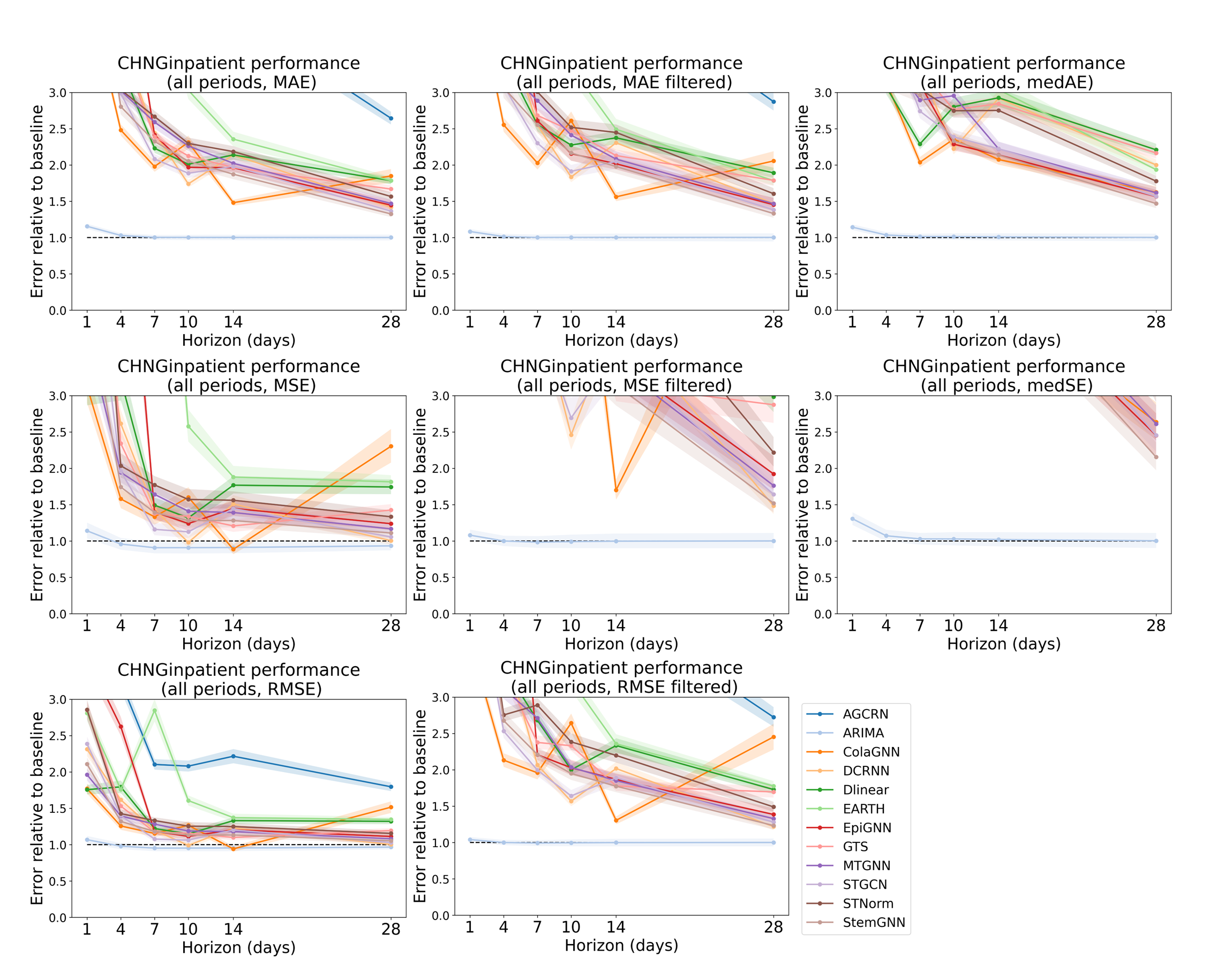}
    \caption{ Error of each method at multiple horizons in CHNGinpatient by each metric. 95\% CI's are calculated for each plot by bootstrapping across months.
    }
\end{figure}
\newpage

\begin{figure}[htb!]
    \centering
    \includegraphics[width=1.0\linewidth]{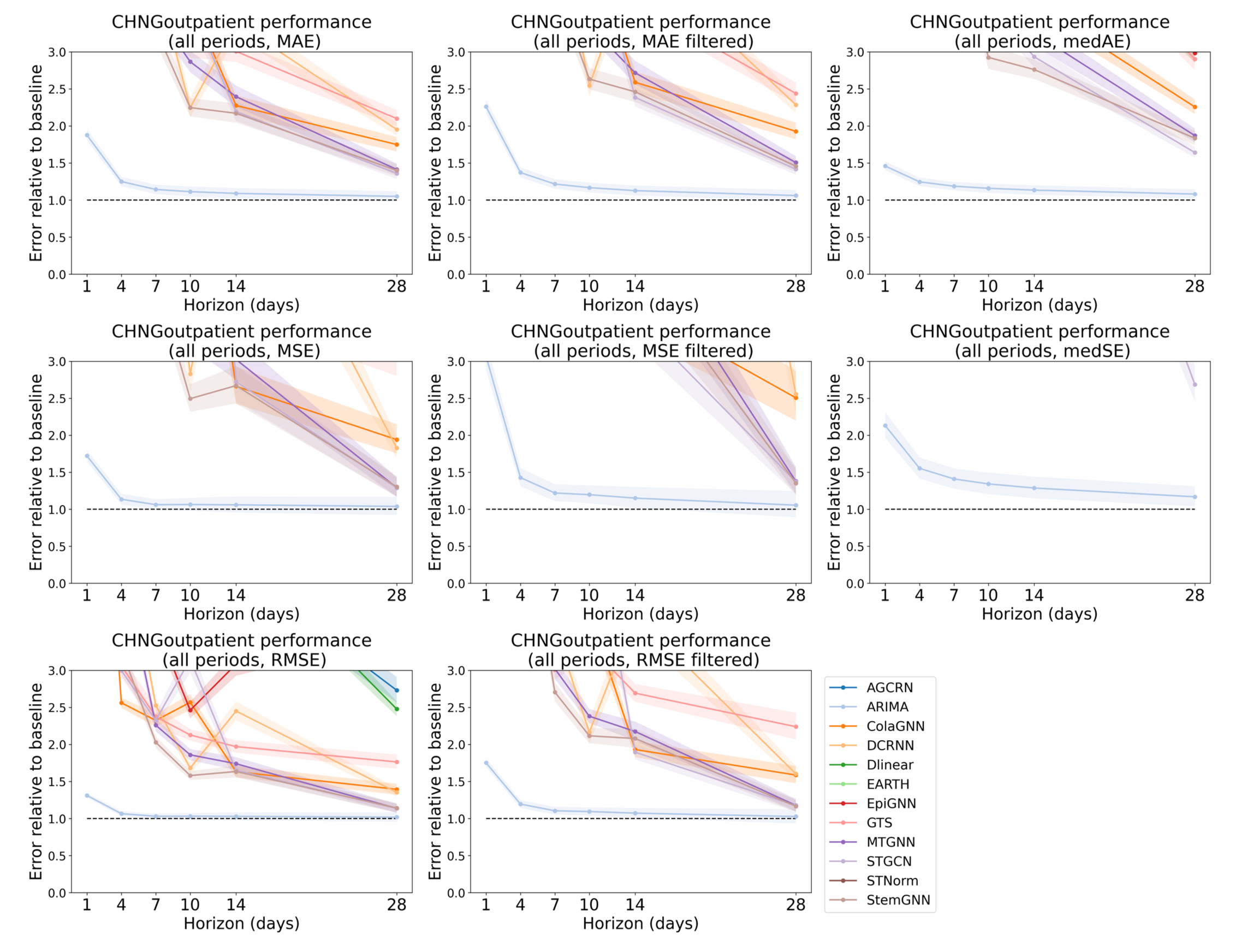}
    \caption{ Error of each method at multiple horizons in CHNGoutpatient by each metric. 95\% CI's are calculated for each plot by bootstrapping across months.
    }
\end{figure}
\newpage

\begin{figure}[htb!]
    \centering
    \includegraphics[width=1.0\linewidth]{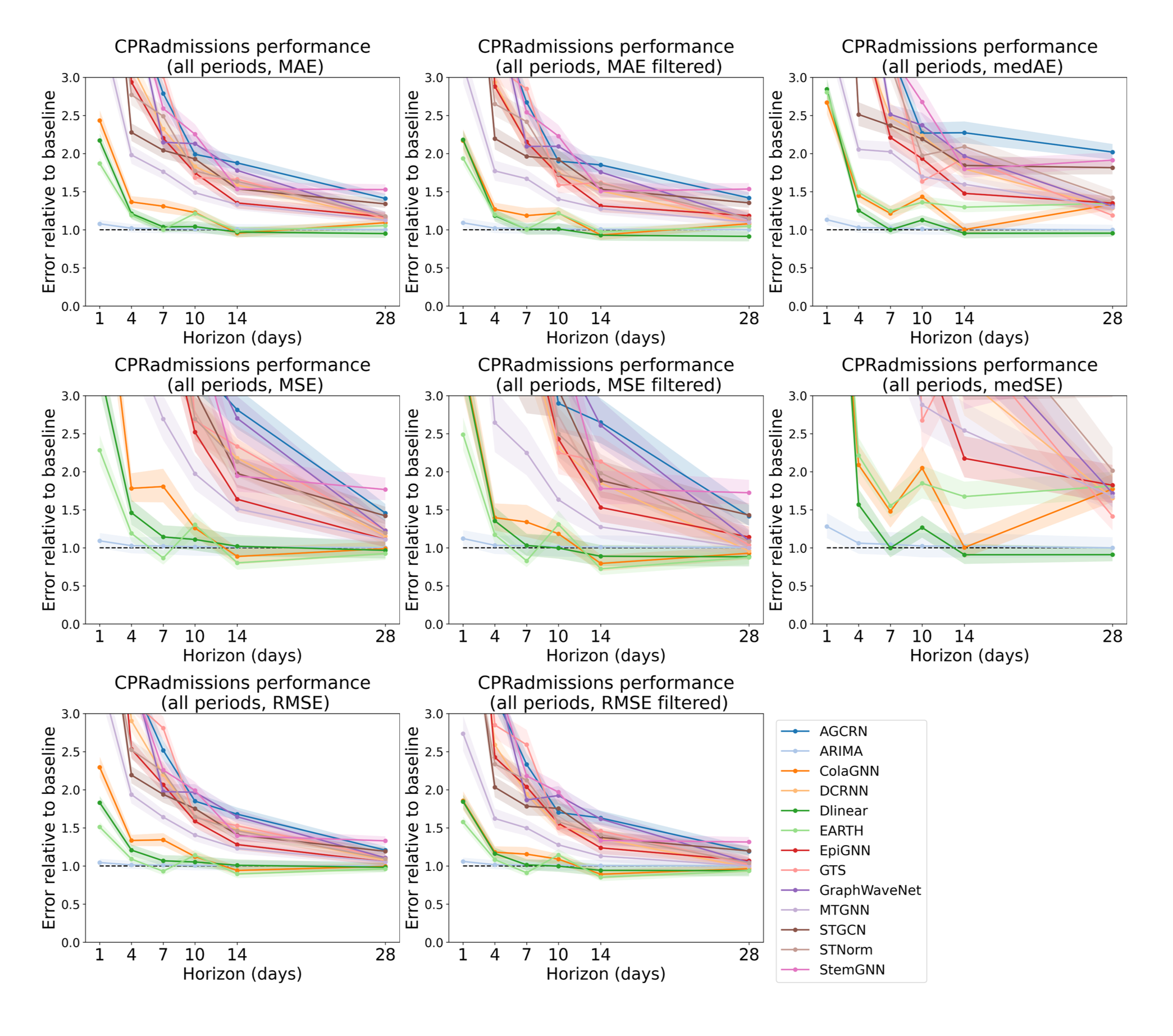}
    \caption{ Error of each method at multiple horizons in CPRadmissions by each metric. 95\% CI's are calculated for each plot by bootstrapping across months.
    }
\end{figure}
\newpage

\begin{figure}[htb!]
    \centering
    \includegraphics[width=1.0\linewidth]{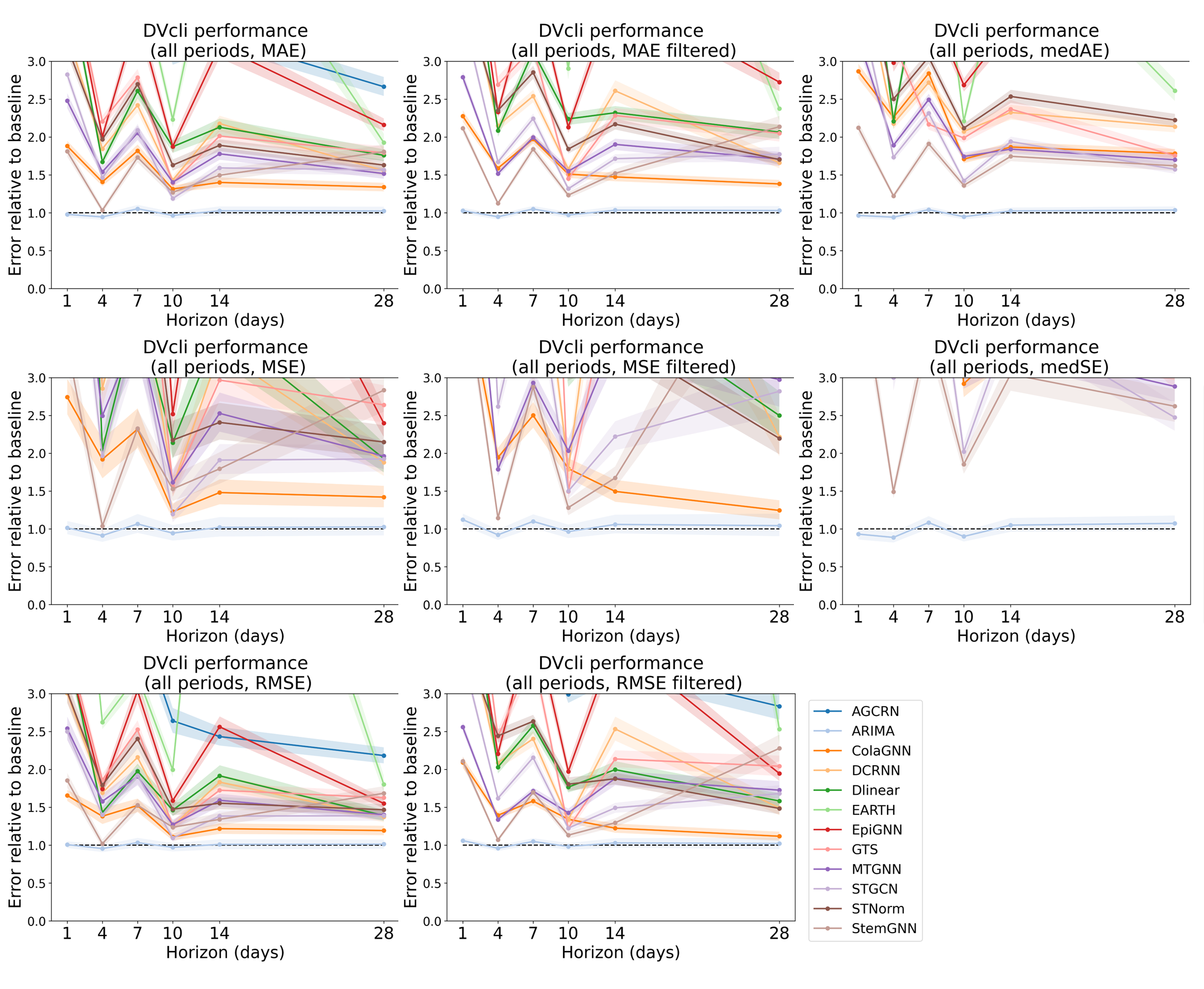}
    \caption{ Error of each method at multiple horizons in DVcli by each metric. 95\% CI's are calculated for each plot by bootstrapping across months.
    }
\end{figure}
\newpage

\begin{figure}[htb!]
    \centering
    \includegraphics[width=1.0\linewidth]{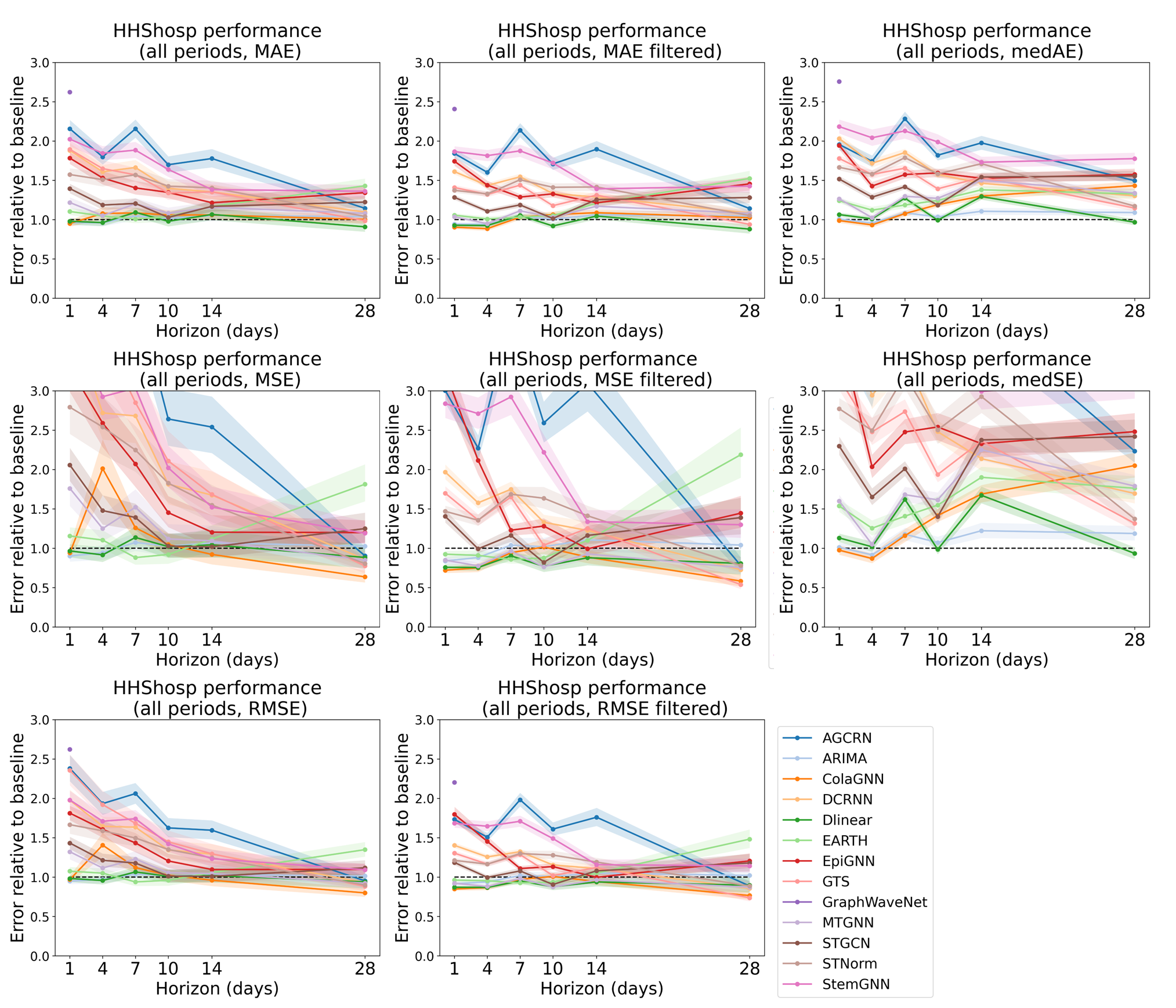}
    \caption{ Error of each method at multiple horizons in HHShosp by each metric. 95\% CI's are calculated for each plot by bootstrapping across months.
    }
\end{figure}
\newpage

\begin{figure}[htb!]
    \centering
    \includegraphics[width=1.0\linewidth]{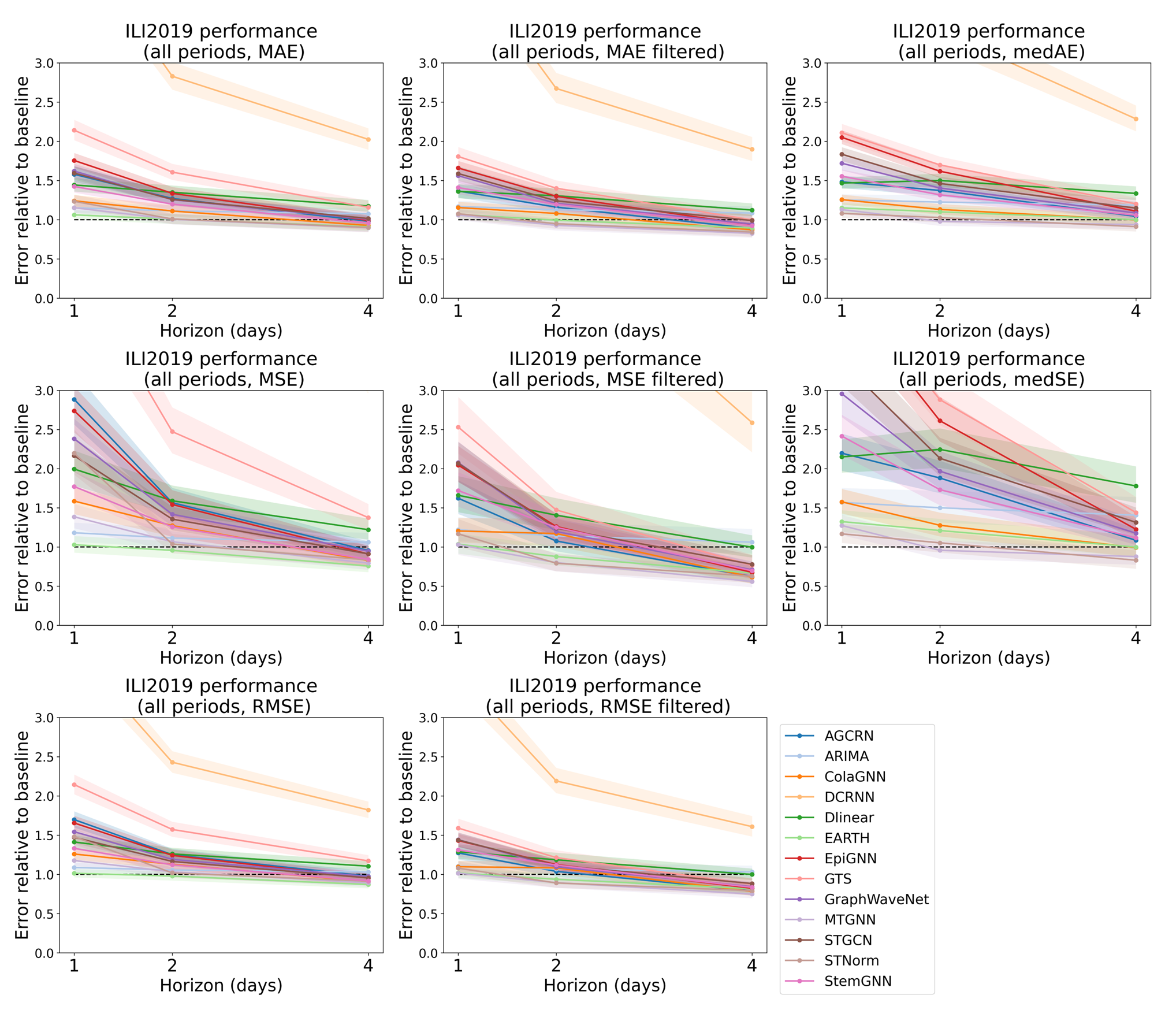}
    \caption{ Error of each method at multiple horizons in ILINet2019 by each metric. 95\% CI's are calculated for each plot by bootstrapping across months.
    }
\end{figure}
\newpage

\begin{figure}[htb!]
    \centering
    \includegraphics[width=1.0\linewidth]{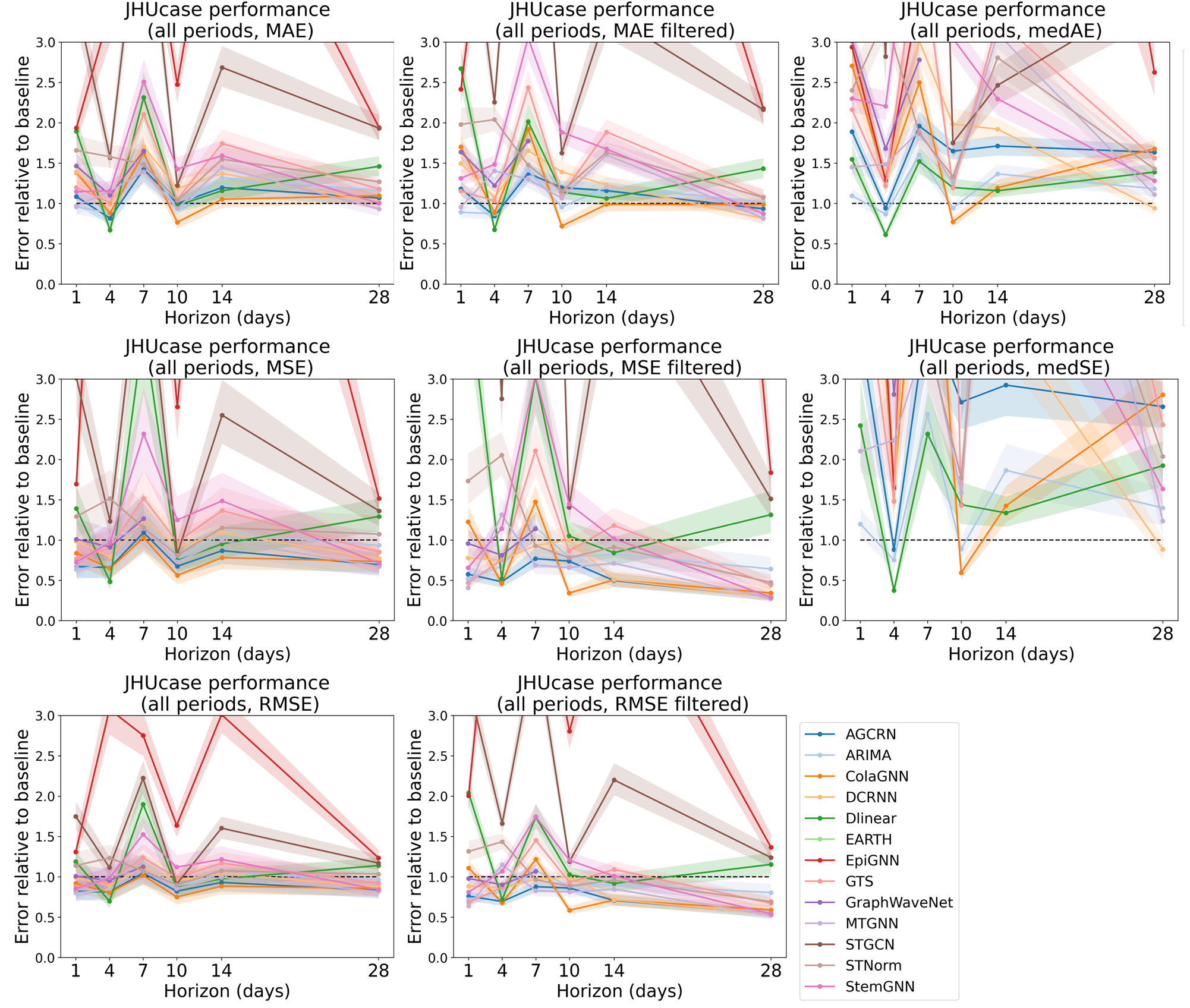}
    \caption{ Error of each method at multiple horizons in JHUcase by each metric. 95\% CI's are calculated for each plot by bootstrapping across months.
    }
\end{figure}

\newpage
\section{Supporting information for Figure 2}
\label{app:fig2}

\begin{figure}[htb!]
    \centering
    \includegraphics[width=1.0\linewidth]{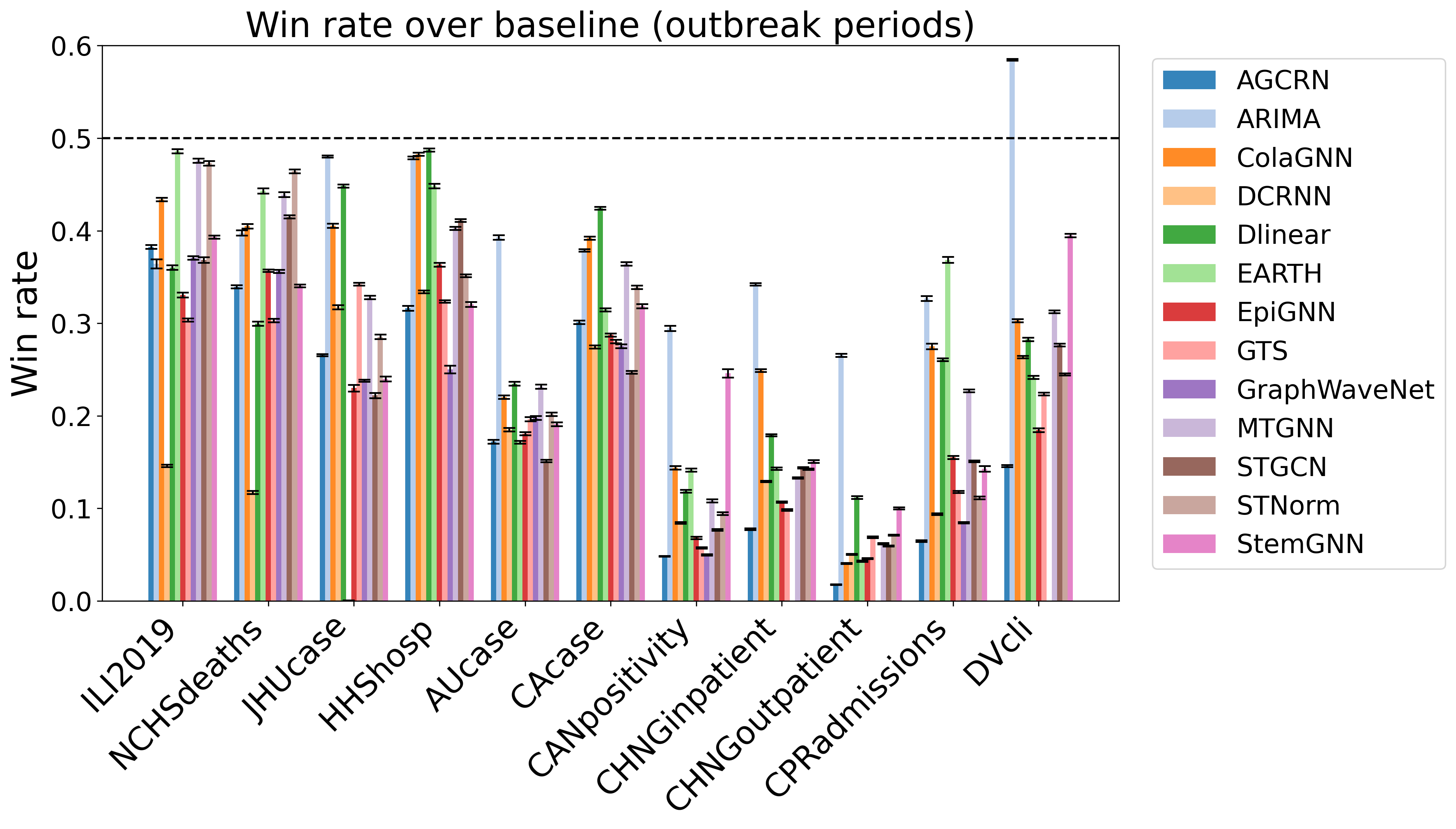}
    \caption{  Win rate of each method versus naive baseline (higher is better), dashed line at 0.5 indicates winning 50\% of the time.
    }
\end{figure}

\newpage
\section{Supporting information for Figure 3}
\label{app:fig3}

\begin{figure}[htb!]
    \centering
    \includegraphics[width=1.0\linewidth]{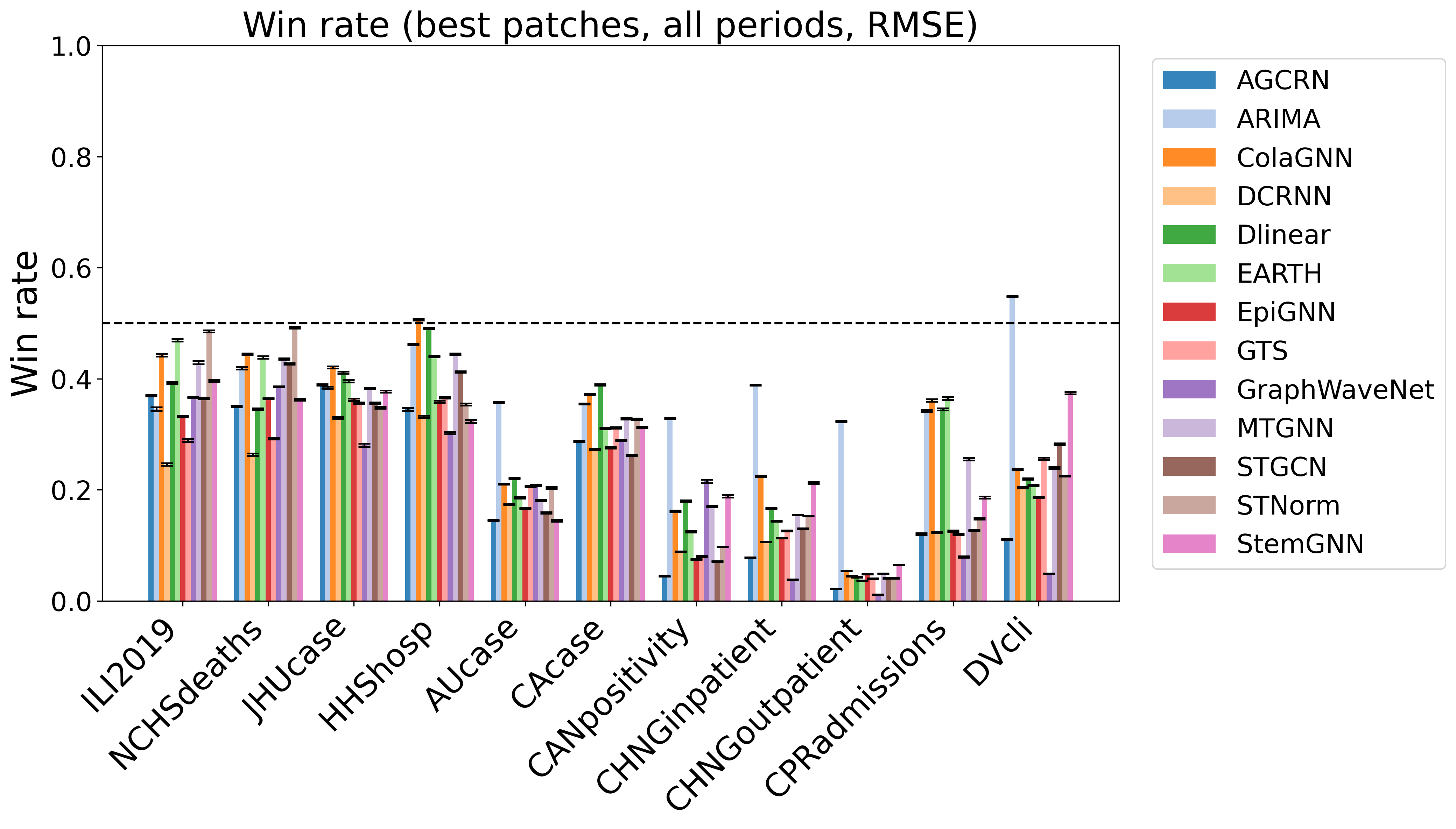}
    \caption{  Win rate of each method versus naive baseline (higher is better), dashed line at 0.5 indicates winning 50\% of the time.
    }
\end{figure}

\begin{figure}[htb!]
    \centering
    \includegraphics[width=1.0\linewidth]{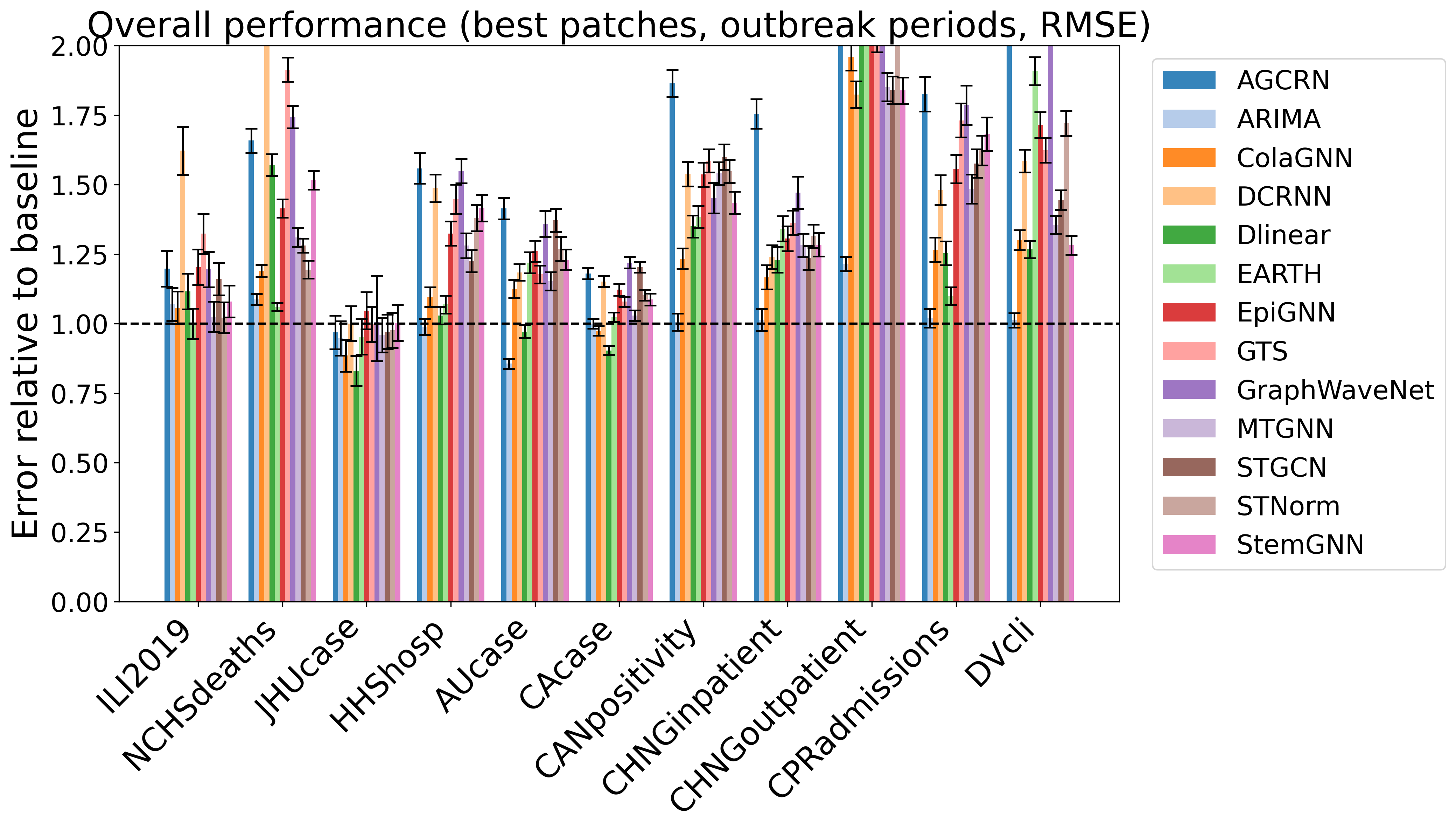}
    \caption{ Error of each method. 95\% CI's are calculated for each plot by bootstrapping across months and meta analyzing across horizons.
    }
\end{figure}

\newpage
\section{Supporting information for Figure 4}
\label{app:fig4}

\begin{figure}[htb!]
    \centering
    \includegraphics[width=1.0\linewidth]{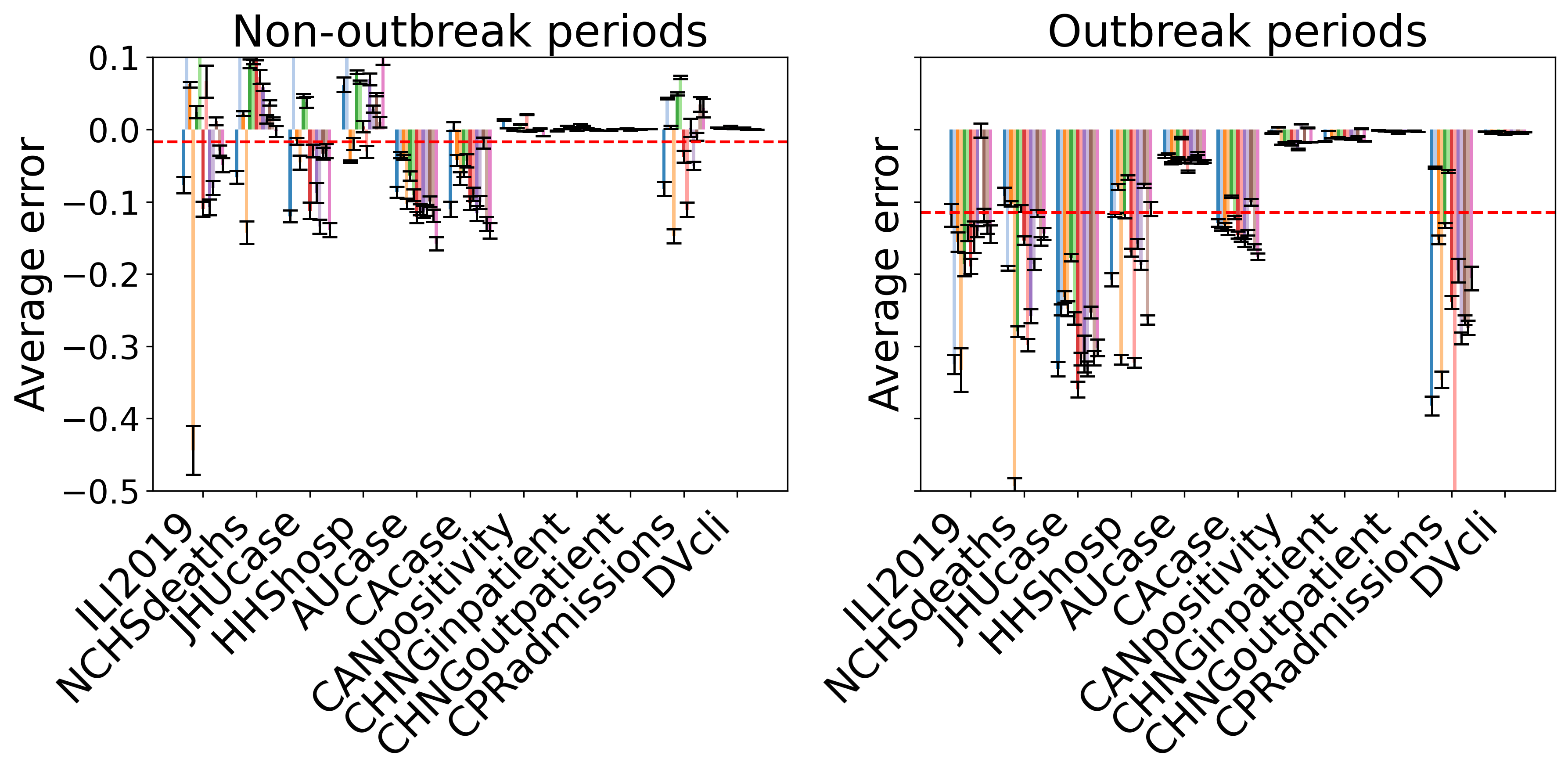}
    \caption{Average error of each method in each dataset for non-outbreaks and outbreaks. Dashed red line indicates overall averge.
    }
\end{figure}

\begin{figure}[htb!]
    \centering
    \includegraphics[width=1.0\linewidth]{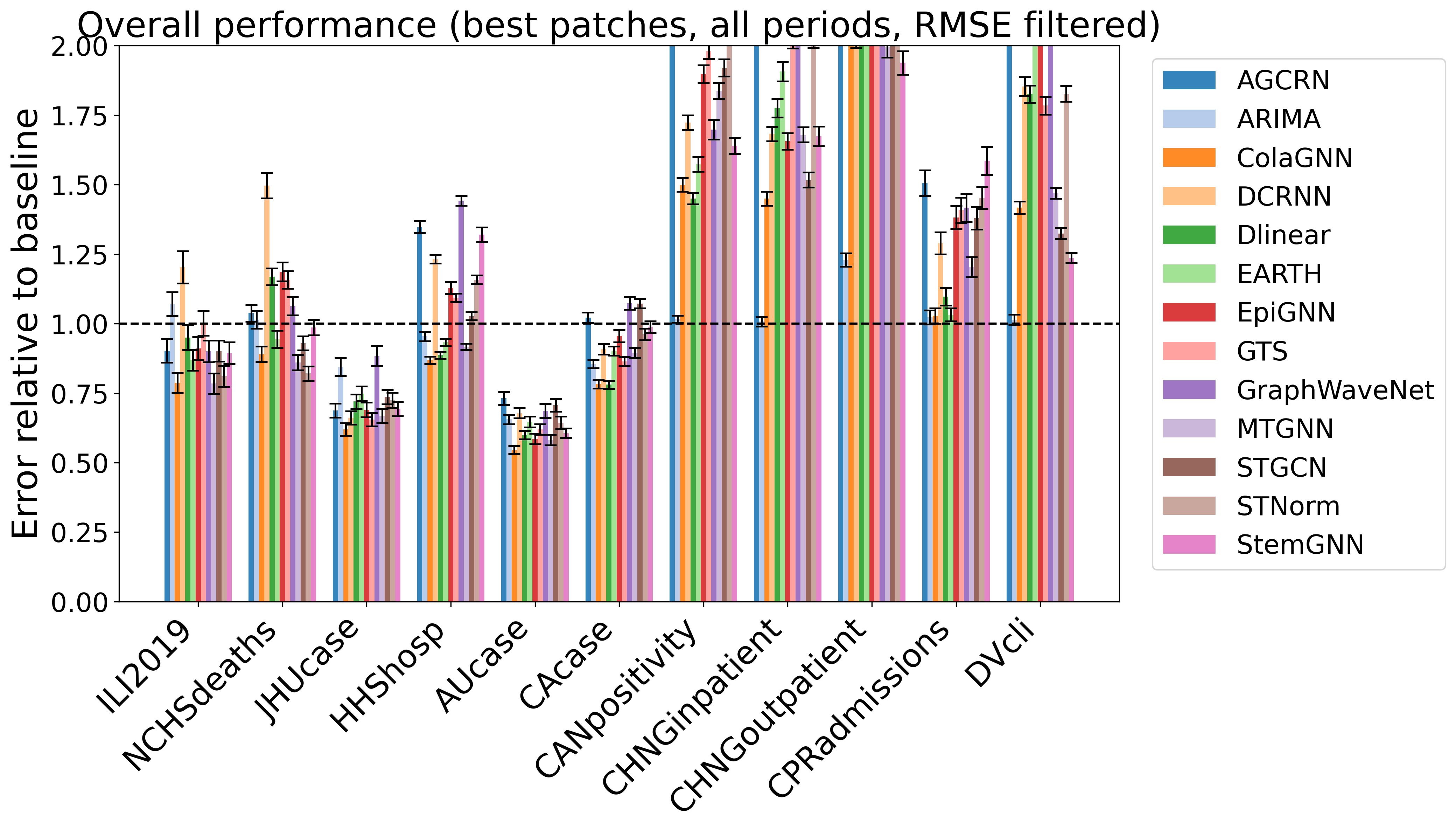}
    \caption{ Error of each method. 95\% CI's are calculated for each plot by bootstrapping across months and meta analyzing across horizons.
    }
\end{figure}

\newpage
\section{Supporting information for Figure 5}
\label{app:fig5}

\begin{figure}[htb!]
    \centering
    \includegraphics[width=1.0\linewidth]{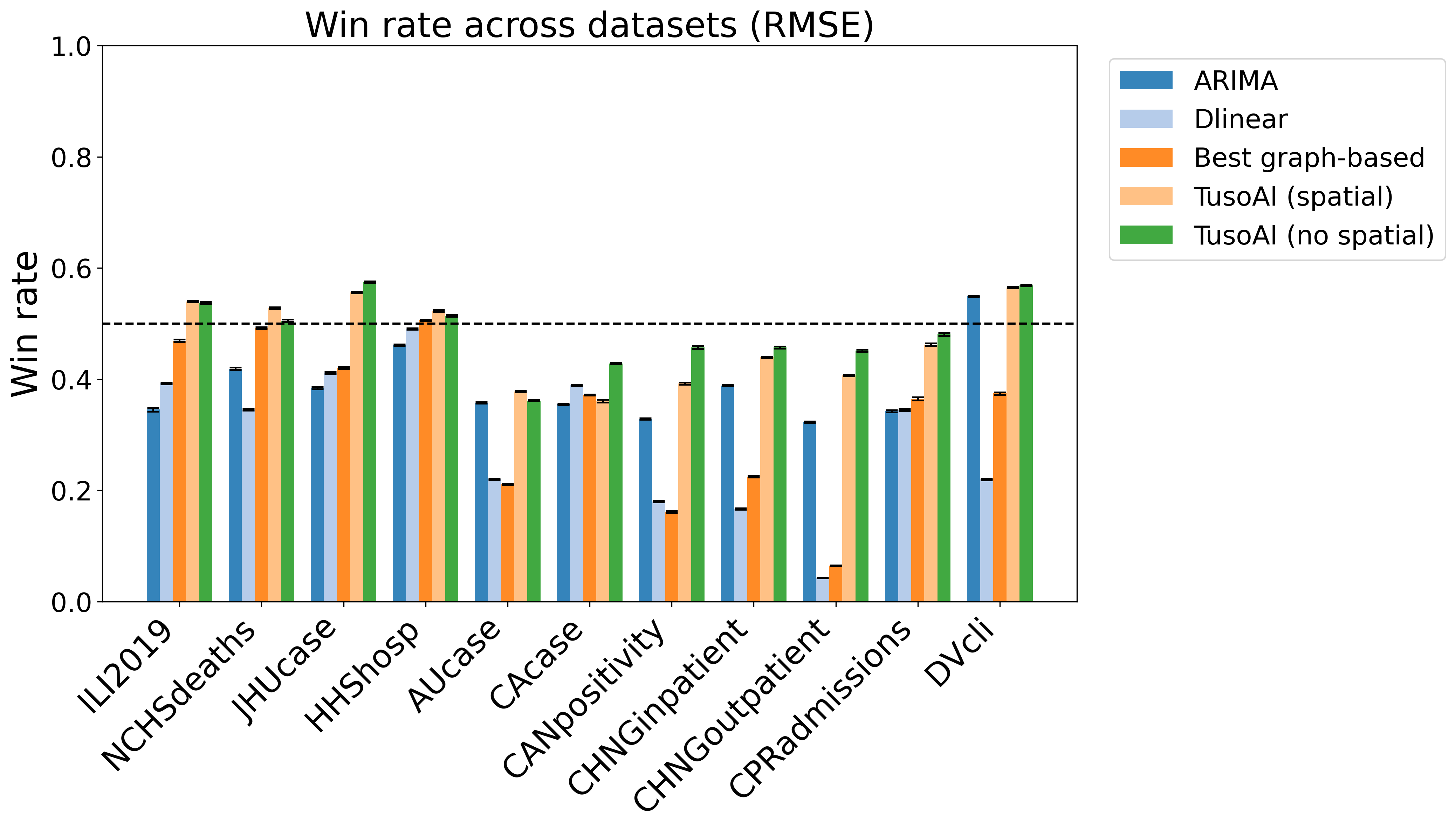}
    \caption{  Win rate of each method versus naive baseline (higher is better), dashed line at 0.5 indicates winning 50\% of the time.
    }
\end{figure}

\newpage
\section*{NeurIPS Paper Checklist}

The checklist is designed to encourage best practices for responsible machine learning research, addressing issues of reproducibility, transparency, research ethics, and societal impact. Do not remove the checklist: {\bf The papers not including the checklist will be desk rejected.} The checklist should follow the references and follow the (optional) supplemental material.  The checklist does NOT count towards the page
limit. 

Please read the checklist guidelines carefully for information on how to answer these questions. For each question in the checklist:
\begin{itemize}
    \item You should answer \answerYes{}, \answerNo{}, or \answerNA{}.
    \item \answerNA{} means either that the question is Not Applicable for that particular paper or the relevant information is Not Available.
    \item Please provide a short (1--2 sentence) justification right after your answer (even for \answerNA). 
\end{itemize}

{\bf The checklist answers are an integral part of your paper submission.} They are visible to the reviewers, area chairs, senior area chairs, and ethics reviewers. You will also be asked to include it (after eventual revisions) with the final version of your paper, and its final version will be published with the paper.

The reviewers of your paper will be asked to use the checklist as one of the factors in their evaluation. While \answerYes{} is generally preferable to \answerNo{}, it is perfectly acceptable to answer \answerNo{} provided a proper justification is given (e.g., error bars are not reported because it would be too computationally expensive'' or ``we were unable to find the license for the dataset we used''). In general, answering \answerNo{} or \answerNA{} is not grounds for rejection. While the questions are phrased in a binary way, we acknowledge that the true answer is often more nuanced, so please just use your best judgment and write a justification to elaborate. All supporting evidence can appear either in the main paper or the supplemental material, provided in appendix. If you answer \answerYes{} to a question, in the justification please point to the section(s) where related material for the question can be found.

IMPORTANT, please:
\begin{itemize}
    \item {\bf Delete this instruction block, but keep the section heading ``NeurIPS Paper Checklist"},
    \item  {\bf Keep the checklist subsection headings, questions/answers and guidelines below.}
    \item {\bf Do not modify the questions and only use the provided macros for your answers}.
\end{itemize}


\begin{enumerate}

\item {\bf Claims}
    \item[] Question: Do the main claims made in the abstract and introduction accurately reflect the paper's contributions and scope?
    \item[] Answer: \answerYes{} 
    \item[] Justification: Each conclusion refers to a corresponding results subsection where we do detailed analysis. The benchmark code is indeed released and ready to use.

\item {\bf Limitations}
    \item[] Question: Does the paper discuss the limitations of the work performed by the authors?
    \item[] Answer: \answerYes{} 
    \item[] Justification: We have an explicit discussion section where we talk about limitations.

\item {\bf Theory assumptions and proofs}
    \item[] Question: For each theoretical result, does the paper provide the full set of assumptions and a complete (and correct) proof?
    \item[] Answer: \answerNA{} 
    \item[] Justification: We have no theoretical results to report.

    \item {\bf Experimental result reproducibility}
    \item[] Question: Does the paper fully disclose all the information needed to reproduce the main experimental results of the paper to the extent that it affects the main claims and/or conclusions of the paper (regardless of whether the code and data are provided or not)?
    \item[] Answer: \answerYes{} 
    \item[] Justification: Yes, our benchmark is fully downloadable and reproducible, and all analysis scripts are uploaded. We document and cite all method and data as well.

\item {\bf Open access to data and code}
    \item[] Question: Does the paper provide open access to the data and code, with sufficient instructions to faithfully reproduce the main experimental results, as described in supplemental material?
    \item[] Answer: \answerYes{} 
    \item[] Justification: Yes, that is a central contribution of our paper.

\item {\bf Experimental setting/details}
    \item[] Question: Does the paper specify all the training and test details (e.g., data splits, hyperparameters, how they were chosen, type of optimizer) necessary to understand the results?
    \item[] Answer: \answerYes{} 
    \item[] Justification: Yes, these are reported both in the code and supplementary material.

\item {\bf Experiment statistical significance}
    \item[] Question: Does the paper report error bars suitably and correctly defined or other appropriate information about the statistical significance of the experiments?
    \item[] Answer: \answerYes{} 
    \item[] Justification: Yes, all plots are supported by appropriate and documented confidence intervals.

\item {\bf Experiments compute resources}
    \item[] Question: For each experiment, does the paper provide sufficient information on the computer resources (type of compute workers, memory, time of execution) needed to reproduce the experiments?
    \item[] Answer: \answerYes{} 
    \item[] Justification: Yes, we document this in the Appendix.
    
\item {\bf Code of ethics}
    \item[] Question: Does the research conducted in the paper conform, in every respect, with the NeurIPS Code of Ethics \url{https://neurips.cc/public/EthicsGuidelines}?
    \item[] Answer: \answerYes{} 
    \item[] Justification: We do not violate any ethical code with this work.

\item {\bf Broader impacts}
    \item[] Question: Does the paper discuss both potential positive societal impacts and negative societal impacts of the work performed?
    \item[] Answer: \answerYes{} 
    \item[] Justification: Yes, throughout the abstract, intro, and discussion we motivate why this task is necessary and how our work will mak epidemic forecasting more reliable.
    
\item {\bf Safeguards}
    \item[] Question: Does the paper describe safeguards that have been put in place for responsible release of data or models that have a high risk for misuse (e.g., pre-trained language models, image generators, or scraped datasets)?
    \item[] Answer: \answerNA{} 
    \item[] Justification: We have no such potential for misuse.

\item {\bf Licenses for existing assets}
    \item[] Question: Are the creators or original owners of assets (e.g., code, data, models), used in the paper, properly credited and are the license and terms of use explicitly mentioned and properly respected?
    \item[] Answer: \answerYes{} 
    \item[] Justification: All data providers and model creators are appropriately cited.

\item {\bf New assets}
    \item[] Question: Are new assets introduced in the paper well documented and is the documentation provided alongside the assets?
    \item[] Answer: \answerYes{} 
    \item[] Justification: Yes, we document all datasets and how they were gathered and preprocessed.

\item {\bf Crowdsourcing and research with human subjects}
    \item[] Question: For crowdsourcing experiments and research with human subjects, does the paper include the full text of instructions given to participants and screenshots, if applicable, as well as details about compensation (if any)? 
    \item[] Answer: \answerNA{} 
    \item[] Justification: We do not do crowdsourcing.

\item {\bf Institutional review board (IRB) approvals or equivalent for research with human subjects}
    \item[] Question: Does the paper describe potential risks incurred by study participants, whether such risks were disclosed to the subjects, and whether Institutional Review Board (IRB) approvals (or an equivalent approval/review based on the requirements of your country or institution) were obtained?
    \item[] Answer: \answerNA{} 
    \item[] Justification: No IRB approval was needed.

\item {\bf Declaration of LLM usage}
    \item[] Question: Does the paper describe the usage of LLMs if it is an important, original, or non-standard component of the core methods in this research? Note that if the LLM is used only for writing, editing, or formatting purposes and does \emph{not} impact the core methodology, scientific rigor, or originality of the research, declaration is not required.
    \item[] Answer: \answerYes{} 
    \item[] Justification: Yes, we run agentic framework TusoAI in a subsection and report how it is used in the Appendix.

\end{enumerate}

\end{document}